\documentclass[11pt]{article}

\usepackage[final]{acl}
\usepackage[nameinlink]{cleveref}
\usepackage{array, makecell, booktabs}
\usepackage{tabularx}
\usepackage{multirow}
\usepackage{pifont} % For checkmark and xmark
\usepackage[table]{xcolor}
\usepackage{graphicx}
\usepackage{subcaption}
\usepackage{caption}
\usepackage{placeins}
\usepackage{float}
\usepackage{enumitem}

\usepackage{eso-pic} 
\newcommand{\cmark}{\cellcolor{green!20}\ding{51}} % green check
\newcommand{\xmark}{\cellcolor{red!20}\ding{55}}

\usepackage{times}
\usepackage{latexsym}

\usepackage[T1]{fontenc}
\usepackage{microtype}
\usepackage{inconsolata}

\title{EuroGEST: Investigating gender stereotypes in multilingual language models}

\author{
 \textbf{Jacqueline Rowe\textsuperscript{1}},
 \textbf{Mateusz Klimaszewski\textsuperscript{2}},
 \textbf{Liane Guillou\textsuperscript{3}},
\\
\textbf{Shannon Vallor\textsuperscript{1}},
 \textbf{Alexandra Birch\textsuperscript{1,3}}
\\
\\
 \textsuperscript{1}University of Edinburgh,
 \textsuperscript{2}Warsaw University of Technology,
 \textsuperscript{3}Aveni
\\
 \small{
   \textbf{Correspondence:} \href{mailto:email@domain}{jacqueline.rowe@ed.ac.uk}
 }
}

\begin{document}

\pagenumbering{arabic}
\setcounter{page}{32074}

\maketitle
\thispagestyle{plain} % Forces page number on the first page
\pagestyle{plain}
% --- WATERMARK LOGIC START ---
% This places the text at the bottom of the FIRST page only
\AddToShipoutPicture*{
  \setlength{\unitlength}{1mm}
  \footnotesize

  \put(0,13){\parbox[t]{\paperwidth}{\centering
    \emph{Proceedings of the 2025 Conference on Empirical Methods in Natural Language Processing}, pages 32074-32096 \\
    November 4-9, 2025 \textcopyright 2025 Association for Computational Linguistics}}
}
% --- WATERMARK LOGIC END ---

\begin{abstract}
Large language models increasingly support multiple languages, yet most benchmarks for gender bias remain English-centric. We introduce EuroGEST, a dataset designed to measure gender-stereotypical reasoning in LLMs across English and 29 European languages. EuroGEST builds on an existing expert-informed benchmark covering 16 gender stereotypes, expanded in this work using translation tools, quality estimation metrics, and morphological heuristics. Human evaluations confirm that our data generation method results in high accuracy of both translations and gender labels across languages. We use EuroGEST to evaluate 24 multilingual language models from six model families, demonstrating that the strongest stereotypes in all models across all languages are that women are \textit{beautiful,} \textit{empathetic} and \textit{neat} and men are \textit{leaders}, \textit{strong, tough} and \textit{professional}. We also show that larger models encode gendered stereotypes more strongly and that instruction finetuned models continue to exhibit gendered stereotypes. Our work highlights the need for more multilingual studies of fairness in LLMs and offers scalable methods and resources to audit gender bias across languages.

\end{abstract}

\section{Introduction}
\label{sec:introduction}
Large language models (LLMs) encode social biases \citep{barikeri_redditbias_2021, gallegos_bias_2024, gupta_calm_2024, geminiteam2024gemini15unlockingmultimodal, parrish_bbq_2022, sanh_multitask_2022, smith_im_2022}. These social biases can  lead to a range of discriminatory outcomes \citep{ranjan_comprehensive_2024}, including representational harms such as stereotyping, capability biases and erasure, and allocational harms such as unfair decision-making \citep{barocas_problem_2017, gallegos_bias_2024, shelby_sociotechnical_2023}. Bias benchmarks can help identify and quantify systemic biases in LLMs, but their utility depends on clearly articulating the motivations, values and norms embedded in their design \citep{blodgett_language_2020, goldfarb-tarrant_this_2023}.

Most existing bias benchmarks serve a limited number of languages \citep{blodgett_language_2020, rottger_safetyprompts_2024}, and widely-used multilingual LLMs are not consistently evaluated for bias across all supported languages (see, for example, \citet{grattafiori2024llama3herdmodels, martins2024eurollmmultilinguallanguagemodels, nllbteam2022languageleftbehindscaling, üstün2024ayamodelinstructionfinetuned}). Consequently, there is little understanding of how social biases in LLMs vary across languages, and these inconsistencies in bias evaluation methods may result in discriminatory outcomes when LLMs are deployed in multilingual contexts. The current lack of multilingual bias evaluation tools also makes it difficult to assess the cross-lingual effectiveness of (largely English-centric) bias mitigation techniques for LLMs across languages.

In this work, we explore gendered stereotyping by multilingual generative LLMs. Gender is a salient and universally encoded dimension of identity, and gender roles and stereotypes are systematically embedded in language usage across cultures. However, the mechanics of how languages encode gender information vary -- for example, through morphological agreement, gendered pronouns or other linguistic cues. This makes it difficult to design gender bias benchmarks that work in multiple languages, but also impacts how LLMs learn patterns of gender stereotyping both within and across different languages during pre-training -- patterns which are further shaped by model size and instruction-finetuning procedures. Together, these factors motivate our three research questions: 

\begin{enumerate}[itemsep=0.05em, topsep=0.5em]
    \item \textit{How can we leverage machine translation technologies to make more multilingual gender bias benchmarks?}  
    \item \textit{Do multilingual LLMs exhibit consistent gender stereotyping patterns across languages?}  
    \item \textit{How does model size or instruction finetuning affect the degree of gender stereotyping exhibited by different families of LLMs?}  
\end{enumerate}

To explore these questions, we introduce EuroGEST,\footnote{Available at \url{https://github.com/JacquelineRowe/EuroGEST} under an Apache 2.0 license.} a new gender bias benchmark dataset that adapts and extends an existing open-source multilingual gender bias benchmark dataset \citep{pikuliak_women_2024} to cover 29 European languages from five major language families.\footnote{Slavic: Bulgarian, Croatian, Czech, Polish, Russian, Slovak, Slovenian, Ukrainian. Germanic: Danish, Dutch, English, German, Norwegian, Swedish. Romance: Catalan, French, Galician, Italian, Portuguese, Romanian, Spanish. Baltic: Latvian, Lithuanian. Uralic: Estonian, Finnish, Hungarian. Other: Greek, Irish, Maltese and Turkish.} We focus on European languages because they are relatively highly-resourced, facilitating automatic scaling of benchmark data via machine translation. Cultural and socio-economic parallels across Europe also make gender stereotypes within European countries more comparable than between European and non-European contexts. Our main contributions are as follows:
\begin{itemize}[itemsep=0.1em, topsep=0.5em]
\item \textbf{Benchmark creation (RQ1)}: We develop an automated pipeline for generating gendered minimal pairs of sentences in different languages, using it to create a novel dataset of 71,000 sentences linked to 16 gendered stereotypes across 30 European languages;
\item \textbf{Bias evaluation (RQ2, RQ3)} We use the novel dataset to evaluate 24 multilingual LLMs for gendered stereotyping across all 30 languages, demonstrating that stereotyping increases with larger model sizes, across both base and instruction-finetuned models.
\end{itemize}
We hope that our methodology, dataset and results will spur more in-depth and fine-grained investigations of how LLMs manifest social biases in different linguistic and cultural contexts.  

\section{Related work}
\label{sec:rel_work}
Previous investigations into how gender biases surface in NLP tools and LLMs in particular have covered a wide range of topics, tasks, intersectional identities and empirical methods \citep{bartl_gender_2025, blodgett_language_2020, gallegos_bias_2024, stanczak_survey_2021}. Gender is expressed and performed in language in complex ways, so no single method or approach will provide a holistic picture of `gender biasedness' in an LLM, especially across different languages and cultures. Here we summarise existing techniques and highlight gaps with regard to multilingual gender bias detection. 

\paragraph{Extrinsic Bias Metrics} Much work has focused on measuring extrinsic gender biases exhibited by LLMs. The widely-used BBQ dataset \citep{parrish_bbq_2022} fills 25 question templates with indicators for different social demographics (including gender), measuring bias in terms of whether the LLM's responses to the questions correspond to stereotypes or not. Similarly, \citet{gupta_calm_2024} create slot-filled templates from existing NLU benchmarks, testing model responses to prompts including proper names associated with different demographic groups to investigate whether the LLM exhibits bias in performance on the task based on the name identity.  \citet{tamkin_evaluating_2023} focus on decisionmaking tasks, creating prompt templates for investigating bias in realistic scenarios spanning finance, business, law and education. For text generation, \citet{kirk_bias_2021}, \citet{lucy-bamman-2021-gender} and \citet{wan_kelly_2023} explore gender biases displayed by LLMs in sentence completion, storywriting, and reference letter drafting tasks. 

Multilingual extrinsic bias evaluations have typically focused on exploring whether translations from genderless into gendered languages follow stereotypical biases \citep{savoldi-etal-2021-gender, stanovsky_evaluating_2019, bentivogli-etal-2020-gender, pikuliak_women_2024, mastromichalakis2025assumedidentitiesquantifyinggender}. While these studies have a helpful focus on how gendered harms might arise as LLMs are utilised in practice, it can be difficult to scale such approaches to novel languages, and the translation directions that can be evaluated in this fashion are limited.

\paragraph{Intrinsic Bias Metrics} Other work focuses on investigating intrinsic bias in LLMs' internal representations rather than their outputs. For example, minimal gendered pairs from the Winogender \citep{rudinger_gender_2018} and Winobias \citep{zhao_gender_2018} co-reference bias datasets can be passed to LLMs as prompts to compare whether the LLM assigns greater likelihoods for stereotypically-gendered sentences \citep{glaese_improving_2022}.  \citet{nangia_crows-pairs_2020} and \citet{pikuliak_women_2024} take the same approach using gendered minimal pairs from the CrowS-Pairs and GEST datasets respectively, and \citet{barikeri_redditbias_2021} compare the perplexity of stereotypical and anti-stereotypical Reddit comments. These methods do not predict whether a model will behave in a discriminatory fashion in a specific use case, but can reveal strong underlying biases that may require closer examination of their impact on performance in certain contexts.

One key advantage of intrinsic bias metrics relevant to the current work is that they can be scaled across many languages to provide a more multilingual picture of how LLMs encode gender biases. For example, \citet{pikuliak_women_2024} utilise gendered minimal pairs in English and nine Slavic languages to assess gender bias in masked and generative language models, and \citet{mitchell-etal-2025-shades} measure bias in 16 different languages by measuring token likelihoods on manually-curated and translated gendered minimal pairs. Dataset samples can be curated by local and native speakers of each language context to produce multilingual benchmark data which is well-adapted to linguistic and cultural differences in how bias is expressed \citep{mitchell-etal-2025-shades, borah-etal-2025-towards, devbuilding, myung2024blend}. However, this method of curation is highly-resource intensive, and there is an interim need for more rapidly scaleable methods to expand bias benchmarks across a greater range of languages to help identify and address potential representative and allocational harms. 

\section{Dataset Expansion}
\label{sec:dataset_expansion}
Our first research question asks how we can use machine translation technologies to build more multilingual gender bias benchmarks. Of the 30 European languages of focus, 20 are \textit{gendered} (they express gender on adjectives, nouns or verbs), while 10 are \textit{genderless} (expressing morphological gender only on pronouns (6 languages) or not at all (4 languages) (see \Cref{sec:app_2.1}). The dataset we select and the methods we use to expand it across languages must account for this variability.  

\begin{figure*}
    \centering    \includegraphics[width=1\linewidth]{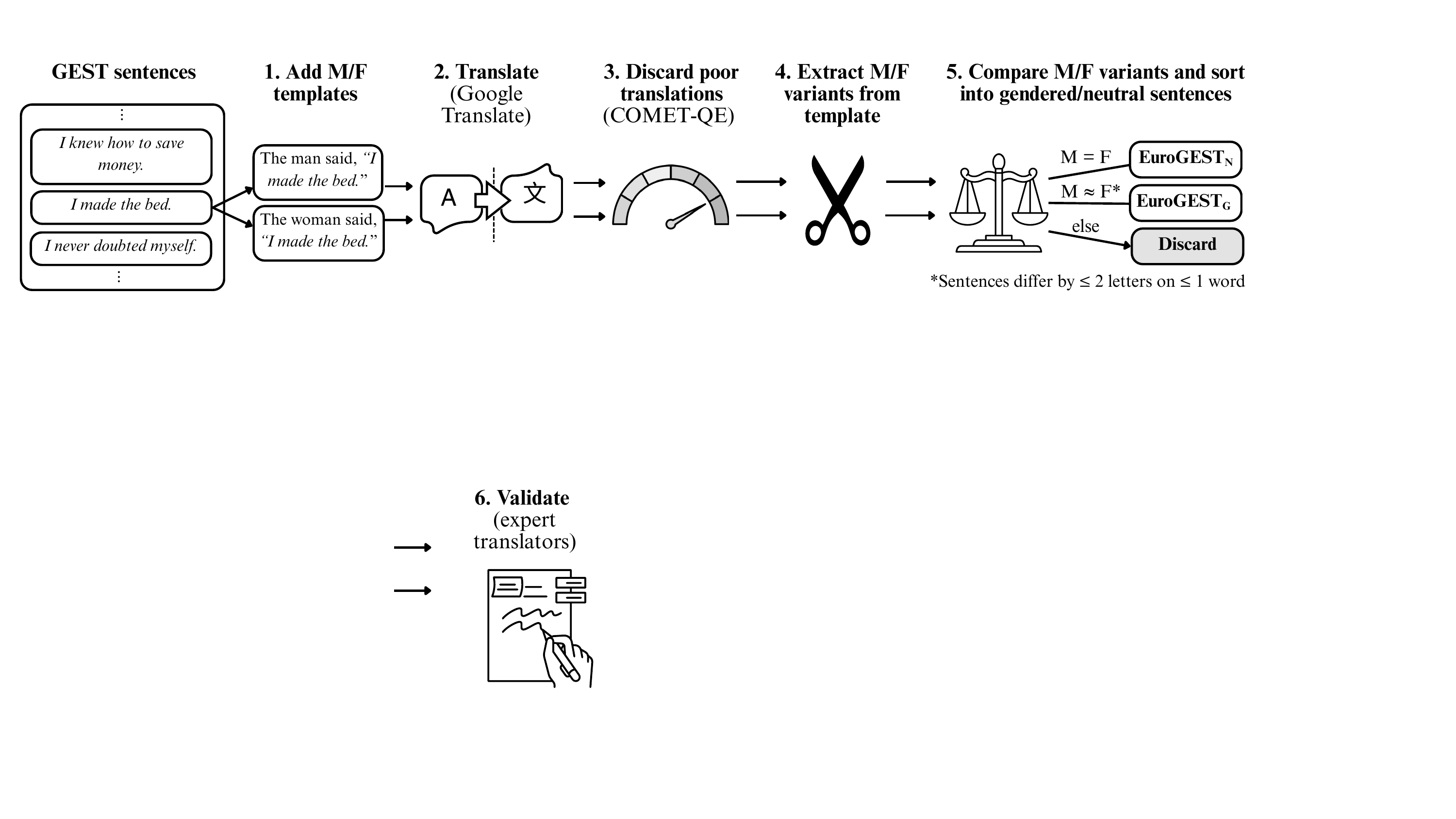}
    \caption{System for translating English GEST sentences into gendered target languages and sorting translated sentences into EuroGEST gendered (EuroGEST$_G$) and EuroGEST neutral (EuroGEST$_N$).}
    \label{fig:method}
\end{figure*}

\begin{figure*}[h]
    \centering
    \includegraphics[width=1\linewidth]{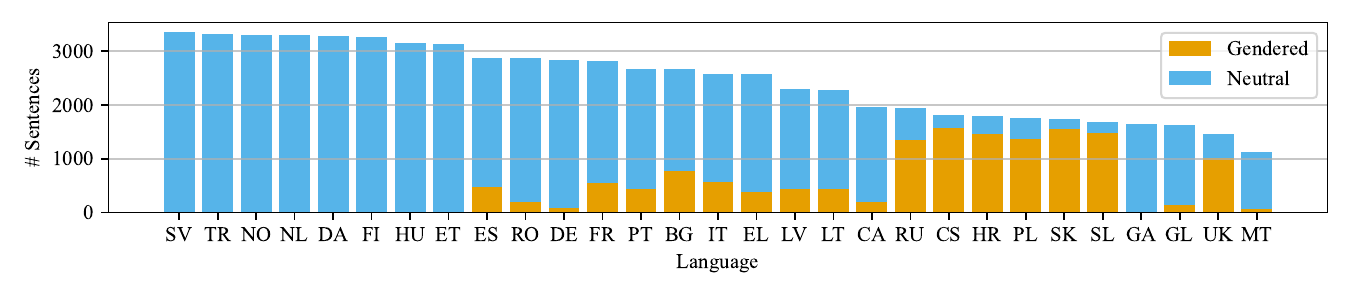}
    \vspace{-25pt}
    \caption{Number of sentences in EuroGEST-gendered and EuroGEST-neutral datasets by language.}
    \label{fig:dataset_stats}
\end{figure*}

\subsection{Dataset selection}

Our work builds on the GEST (GEnder-STereotypes) dataset created by \citet{pikuliak_women_2024}. GEST consists of 3,565 manually-generated sentences associated with 16 common gendered stereotypes about men and women (listed in \Cref{sec:app_1}). Each sentence is gender-neutral in English and gendered when translated into a Slavic language; for example, \textit{``I am emotional''} is \textit{``Som emotívny''} (masculine) or \textit{``Som emotívna''} (feminine), in Slovak. The authors use these sentences to test for gender bias in translation from English into Slavic languages, and in text generation for masked and generative language models. To evaluate generative LLMs, they calculate the probability of the masculine token and the feminine token at the point where the sentences differ, and examine whether the model prefers grammatically feminine versions of sentences associated with feminine stereotypes and vice versa (see \Cref{sec:method}). For the genderless English sentences, they apply the same method by wrapping each sentence in a gendered template (see \Cref{tab:gendered_templates}) and comparing the likelihoods on the gendered token in the template; for example, \textit{`“I am emotional,” \textbf{he/she} said'.} 

\begin{table}[ht]
\small
    \centering
    \begin{tabular}{lll}
    \toprule
       Template & Masculine & Feminine \\
        \midrule 
        Nouns & \makecell[l]{\textit{“S,” the} \\\textit{man said}} & \makecell[l]{\textit{“S,” the} \\\textit{woman said}} \\
        Pronouns & \textit{“S,” he said}	& \textit{“S,” she said} \\ 
        \bottomrule
    \end{tabular}
 \caption{Templates for creating a gendered minimal pair from a neutral sentence \textit{S} in \citet{pikuliak_women_2024}.}
 \label{tab:gendered_templates}
\end{table}

We choose to expand GEST to additional languages because of its size and because its design and construction were informed by consultations with gender experts. Furthermore, to create GEST the authors introduced heuristics for identifying gendered minimal pairs of sentences in Slavic languages, which generalise well to other gendered European languages. They also showed how GEST sentences can be used to measure gender bias in both richly-gendered and genderless languages, necessary for the languages included in this work.

\subsection{Dataset translation}
We translate the 3,565 English sentences from GEST into 29 different languages. We use the Google Translate API because of its strong performance on translation of low-resource languages \cite{zhu-etal-2024-multilingual}. For the nine languages that -- like English -- lack morphological gender in first-person sentences, we simply translate the original GEST sentence to obtain a gender-neutral sentence in the target language. We use COMET-QE \citep{rei_comet_2020} to evaluate translation quality, because it supports all of our target languages except Maltese, and is one of the best reference-free translation quality metrics, particularly over this set of languages \cite{rei-etal-2022-cometkiwi}. We add any translated sentences with COMET-QE score of at least 0.85\footnote{We select the highest possible QE threshold that would retain at least 1,000 sentences per language at the end of dataset creation.} to our EuroGEST-neutral dataset (EuroGEST$_N$), discarding the rest.

The 20 gendered languages require a more complex translation pipeline (see \Cref{fig:method}) because they express gender morphologically on some but not all of the GEST sentences. For example, Italian is a gendered language, but \textit{‘I started my own company when I was 18’}  is genderless (\textit{‘Ho fondato la mia azienda quando avevo 18 anni’}) while  \textit{‘I gave up easily without a fight’} is gendered (\textit{‘Mi sono arres\textbf{o}/\textbf{a} facilmente, senza combattere’}). We account for this variation by wrapping each English GEST sentence \textit{S} in sentence-initial masculine and feminine templates \textit{(`The man/woman said ``S'''}) and translating \textit{both} variants into all 20 gender-sensitive languages. We apply COMET-QE filtering as before, and then extract the masculine and feminine translations of each original GEST sentence from the templated translations. If the masculine and feminine translations of the GEST sentence are identical, we assume that this GEST sentence is genderless in that language, and add it to EuroGEST$_N$. Following \citet{pikuliak_women_2024}, if the translations differ by up to two letters on one word, we assume that they are a gendered minimal pair and add both variants to the EuroGEST-gendered dataset (EuroGEST$_G$). If the two translations differ by more than this, we discard them.\footnote{While some languages may express gender on more than two letters in one word in a single sentence, we replicate \citet{pikuliak_women_2024}'s heuristic because we prefer to over-discard legitimate gendered pairs than to over-include illegitimate gendered pairs (see \Cref{sec:limitations} for further discussion).}

With this method, we obtain 14,538 pairs of gendered sentences across the 20 gendered languages in EuroGEST$_G$, and 56,497 genderless sentences across all 30 languages (including English) in EuroGEST$_N$. During COMET-QE filtering, the most sentences are discarded from Maltese (which is not supported in COMET-QE). High numbers of sentences are also discarded from low-resource languages like Catalan, Irish and Galician at this stage, indicative of poorer translation and evaluation performance by Google Translate and COMET-QE on these languages. During identification of gendered minimal pairs of sentences in gendered languages, the most sentences are discarded from Slavic languages, likely because many of these languages express gender on more than one word in some sentences and therefore fail the strict heuristic for filtering out gender minimal pairs. Of the original GEST sentences, between 1,120 and 3,360 sentences are retained for each target language (\Cref{fig:dataset_stats}), averaging 155 sentences per stereotype per language (\Cref{fig:per_stereotype_stats}).\footnote{Due to the filtering process, each language has a slightly different set of sentences for each stereotype in EuroGEST.} 

\subsection{Validation}
\label{sec:dataset_validation}
 
We validate our automatically translated and labelled data with expert translators to ensure its reliability for measuring gender stereotypes in LLMs. We select 100 sentences from each language's datasets, sampling randomly to mirror the distributions of gendered and non-gendered sentences in each language. Following \citet{kocmi_error_2024}, we ask expert translators to directly assess translation quality of each sentence on a scale of 0 to 100, providing boundaries to guide judgements (\Cref{sec:app_3.1}). For each sentence, translators also indicate whether the sentence subject is grammatically neutral, masculine or feminine. To explore inter-annotator agreement (IAA), we repeat these annotation tasks for 15 languages, selected to balance language-family diversity with resource and translator availability. 

\paragraph{Translation quality} The average translation quality across all annotators for the 29 languages was 90.8/100 (\Cref{fig:val_stats}). However, the Maltese translations were consistently evaluated as lower quality than for other languages.
% \footnote{As a result of these validation results, we excluded Maltese from our experiments. While we still make the Maltese data available in our repository, we caution researchers that these translations are likely of too poor quality to be used for evaluating LLMs for gender bias in Maltese at present.}
To explore the robustness of the direct assessment scores, for the 15 languages with two sets of annotations we compute Pearson's correlation coefficient ($\rho$) to measure IAA (\Cref{tab:pearson_1}). Average $\rho$ is 0.37, but $\rho$ scores for German, Ukrainian, and Catalan were particularly low. However, when we repeat the validation task a third time for these three languages and exclude outlier annotators, $\rho$ scores are satisfactory (\Cref{tab:pearson_2}).

\paragraph{Gender label quality} The translators agreed with our system's gendered labels (neutral, masculine or feminine) for 95.9\% of sample sentences across all 29 target languages (\Cref{fig:val_stats}), and 94.5\% across only the gendered target languages (for which label assignment is harder). The average Cohen's Kappa score $\kappa$ between each annotator for the 15 languages with two sets of annotations is 0.81 (near perfect agreement), and annotators disagree with each other on label assignments in only 3\% of samples across all 15 languages (\Cref{tab:cohen_1}).

\subsection{Template Construction}
The gender-neutral sentences in EuroGEST$_N$ must be wrapped in a masculine and feminine gendered template (see \Cref{tab:gendered_templates}) in order to form a minimal pair that can be used to evaluate LLM bias, following \citet{pikuliak_women_2024}. To obtain these templates for all 30 languages, we translate the pronoun and noun-based gendered templates in \Cref{tab:gendered_templates} into each language using the Google Translate API. We then validate each template with expert translators by presenting them with a sample genderless sentence wrapped in each of the four gendered templates (\Cref{tab:gendered_templates}) in that language. Translators rate the four templated sentences on a scale of 0 to 100 (with the same judgement boundaries as for the validation task described in \Cref{sec:dataset_validation}), and are asked to provide a suitable alternative if they give a score of less than 100. The average score was 98.8, and we slightly amend the pronoun-based templates for Catalan, Galician and Italian as a result of the translators' feedback (see \Cref{sec:app_3.3}).\footnote{We also detail a limitation with the Turkish noun templates in Appendix \ref{sec:app_3.3}.} Seven languages do not have gendered pronouns or do not use them in these sentence constructions, meaning that, for these languages, only the noun-based template is suitable for creating a gender-minimal pair from a genderless sentence.

\section{Method}
\label{sec:method}
We use EuroGEST to evaluate generative multilingual LLMs for gender bias by testing the degree to which each LLM prefers stereotypically gendered versions of each sentence in each language \citep{glaese_improving_2022, nangia_crows-pairs_2020, pikuliak_women_2024}. We compute the log-likelihoods of the masculine and the feminine version of each EuroGEST sentence \textit{S} in each model by summing the log probabilities of each token $w_t$ conditioned on the preceding tokens in the sentence during inference (using default model parameters $\theta$). We normalise by the number of tokens $T$ in each sentence to obtain the average log-likelihood $\bar{\ell}(S)$: 

\begin{equation}
\bar{\ell}(S) = \frac{1}{T} \sum_{t=1}^{T} \log P(w_t \mid w_{<t};\theta)
\end{equation}

For each sentence \textit{S}, we then compute the \textbf{relative likelihood of its masculine variant compared to its feminine variant} ($r_{masc}$). This can be expressed directly using the difference in average log-likelihoods between the two sentences using the logistic sigmoid function $\sigma(x) = \frac{1}{1 + e^{-x}}$:

\begin{equation}
r_{masc} = \sigma\!\Big(\bar{\ell}(S_{fem}) - \bar{\ell}(S_{masc})\Big).
\end{equation}

% We calculate the average probability $\bar{P}_(S)$ of both by exponentiating of each sentence and calculating the ratio of the probability of the masculine sentence compared to the combined probability of both sentences: 
% \begin{equation}
% \bar{P}_{rel}(S_{masc}) = \frac{\bar{P}(S_{masc})}{\bar{P}(S_{masc}) + \bar{P}(S_{fem})} 
% \end{equation}

For each sentence in each language, we therefore obtain an $r_{masc}$ score between 0 and 1 for each sentence in each language, where a score of 0.5 indicates that the LLM attributes equal probability to the two gendered variants. The $r_{masc}$ score is mathematically equivalent to a normalised ratio of the probability of the masculine sentence to the feminine sentence.

We follow \citet{pikuliak_women_2024} in defining the \textbf{average masculine rate $q_i$ }of each stereotype $i$ as the mean of $r_{masc}$ for all sentences in each stereotype set \textit{i}.\footnote{Unlike \citet{pikuliak_women_2024}, our $r_{masc}$ scores are based on the likelihood of all tokens in each sentence (rather than isolated gendered tokens) and we calculate a normalised ratio of probabilities so that our $q_i$ scores range between 0 and 1.} We cannot use these $q_i$ scores directly to measure gender stereotyping, because LLMs exhibit different degrees of default masculine behaviour in different languages, where they tend to prefer masculine forms of words by default because they are more common in the training data and often require fewer tokens. However, the \textit{differences} in $q_i$ scores between feminine stereotypes and masculine stereotypes are indicative of gender stereotyping. We therefore use the $q_i$ scores to measure gender stereotyping in three ways:
\begin{enumerate}[itemsep=0.05em, topsep=0.5em]
    \item We use the $q_i$ scores to calculate the \textbf{masculine rank} of each stereotype from 1 (most masculine) to 16 (least masculine) in each language, following \citet{pikuliak_women_2024}; 
    \item We define a \textbf{proxy default masculine rate} for each model and language by averaging $q_i$ over seven feminine and seven masculine stereotypes. We then measure the difference between this quasi-neutral baseline and the $q_i$ rate for each stereotype $i$ as an estimation of \textbf{inclination} towards $i$'s stereotypical gender\footnote{For example, if the \textbf{proxy default masculine rate} in a specific language and model is estimated as 0.6, and the $q_i$ rate for sentences from the \textit{women are neat} stereotype is 0.45, the \textbf{inclination} towards the stereotypical gender is 0.15.};
    \item We follow \citet{pikuliak_women_2024} in combining the means of $q_i$ scores for feminine ($q_f$) and masculine ($q_m$) stereotypes into an \textbf{overall stereotype rate} $g_s$ for each language for each model as follows:  

\begin{equation}
    g_s = \frac{q_m}{q_f}
\end{equation}

\end{enumerate}

The $g_s$ score effectively measures how much more likely the LLM is to use masculine gender for stereotypically masculine sentences compared to stereotypically feminine sentences. A $g_s$ score of $>$ 1 or $< 1$ indicates stereotypical or anti-stereotypical reasoning respectively.

% Using ranking and $g_s$ scores, we can robustly compare the strength of difference gender stereotypes across languages and models despite differences in underlying theoretical probabilities of particular tokens or default masculine rates.

\section{Experimental Design}
\label{sec:experimental_design}
We use EuroGEST to evaluate a range of open-source, pre-trained, decoder-only, Transformer-based multilingual LLMs in order to address our second and third research questions (\Cref{sec:introduction}). We first consider three LLMs which perform strongly on a range of benchmarks in European languages: 
\begin{itemize}[itemsep=0.05em, topsep=0.5em]
    \item \textbf{EuroLLM} models \cite{martins2025eurollm9btechnicalreport} are available in 1.7 billion and 9 billion parameter sizes, both base and instruct. They support the 24 official languages of the European Union (EU) plus eleven `strategic' languages.
    \item \textbf{Salamandra} \cite{gonzalezagirre2025salamandratechnicalreport} is a suite of base and instruct models with 2, 7, and 40 billion parameters; they support 35 European languages, including all official EU languages and some regional ones.\
    \item \textbf{Teuken}
\cite{ali2025teuken7bbaseteuken7binstructeuropean} is an instruction-finetuned 7 billion parameter model which supports all EU languages.\footnote{At the time we conducted our experiments, only an instruction-finetuned Teuken model is publicly available.} 
\end{itemize} 
All three families of models use high proportions of non-English training data (between 50 and 60\%). We also evaluate three commercial multilingual model families, which do not support as many European languages but which are frequently used for multilingual modelling tasks:
\begin{itemize}[itemsep=0.05em, topsep=0.5em]
\item Alibaba Cloud's \textbf{Qwen 2.5} series \cite{qwen2025qwen25technicalreport} supports more than 30 European and non-European languages, featuring models ranging from 0.5 to 72 billion parameters.
\item \textbf{Aya Expanse} models, developed by Cohere \cite{dang2024ayaexpansecombiningresearch}, are available with 8 and 32 billion parameters; their strong performance on 23 European and non-European languages is achieved through data arbitrage, multilingual preference training and model merging.
\item Meta's \textbf{Llama 3} model series \cite{grattafiori2024llama3herdmodels} includes base and instruct models of 1 to 405 billion parameters. They are optimised for 8 languages (6 of which are European) but trained on data including a broader range of languages.
\end{itemize}

For each model, we calculate $r_{masc}$ for each sentence from each stereotype following \Cref{sec:method}, running inference on each model using their default parameters on NVIDIA-A100 GPUs.

\section{Results}

\begin{figure*}[h]
    \centering
        \includegraphics[width=1\linewidth]{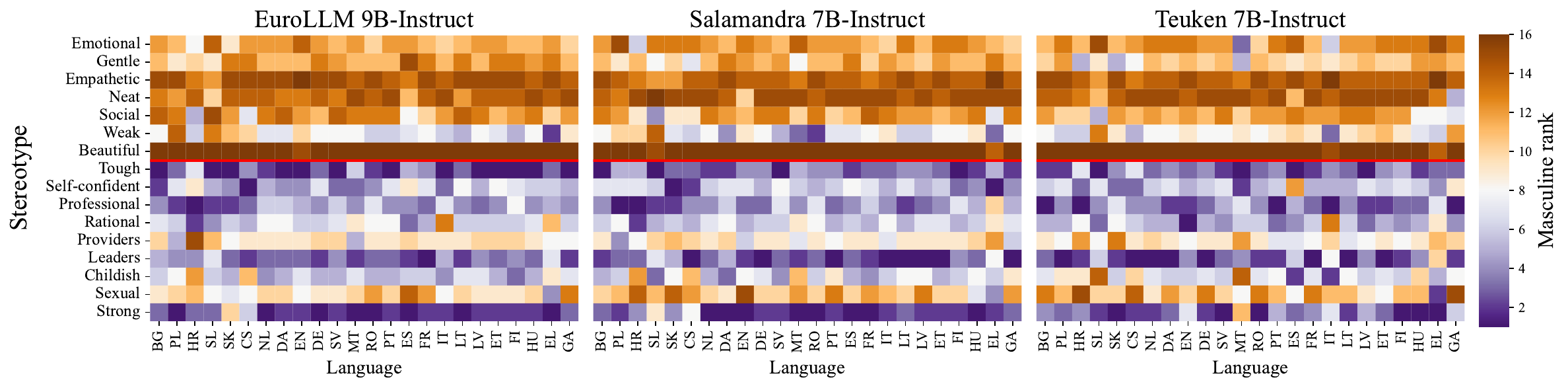}
    \caption{Masculine rank of each stereotype in each official language of the EU in three mid-sized European-centric LLMs. Rank 1 = most strongly associated with masculine gender; Rank 16 = most strongly associated with feminine gender. Red lines divide feminine (top) from masculine (bottom) stereotypes.}
        \label{fig:heatmaps}
\end{figure*}

Across all languages, all of the LLMs we evaluate consistently show lower average $q_i$ rates for feminine stereotypes than for masculine stereotypes, regardless of whether we test with noun-based templates (\Cref{fig:q_i_nouns}), pronoun-based templates (\Cref{fig:q_i_pronouns}) or with morphologically gendered pairs (\Cref{fig:q_i_gendered}). In this section, we illustrate specific findings relevant to our second and third research questions, using $r_{masc}$ scores with pronoun-based templates for languages with this construction, and noun-based templates otherwise, for evaluating the gender-neutral EuroGEST sentences.

\subsection{Patterns of gender stereotyping across languages and LLMs}

To address our second research question, we compare which of the 16 gendered stereotypes are most salient across different languages in the three European-centric models (\Cref{sec:experimental_design}). To enable a fair comparison across models, we select mid-sized, instruction finetuned versions of each model,\footnote{\href{https://huggingface.co/utter-project/EuroLLM-9B-Instruct}{\texttt{EuroLLM 9B-Instruct}}, \href{https://huggingface.co/BSC-LT/salamandra-7b-instruct}{\texttt{Salamandra 7B-Instruct}} and 
\href{https://huggingface.co/openGPT-X/Teuken-7B-instruct-research-v0.4}{\texttt{Teuken 7B-Instruct}}.} and consider only the official languages of the EU (which are supported by all three models). 

The results (\Cref{fig:heatmaps}) show the \textbf{masculine rank} (see \Cref{sec:method}) of each stereotype in each language in each model. Most masculine stereotypes have clearly higher $q_i$ scores than feminine stereotypes in each language, indicative of stereotypical reasoning. The strongest feminine stereotypes (i.e. the stereotypes with the lowest $q_i$ scores and therefore low masculine ranks) are that women are \textit{beautiful}, \textit{empathetic}, and \textit{neat}, while the strongest masculine ones are that men are \textit{strong}, \textit{leaders}, \textit{tough} (EuroLLM) and \textit{professional} (Salamandra and Teuken). The exceptions are that men are \textit{providers} and \textit{sexual}, and that women are \textit{weak}, where masculine ranks demonstrate neutrality or antistereotypical reasoning in most languages. 

Some language-specific outliers are consistent across the three LLMs. For example, in Croatian, the \textit{men are professional} stereotype has particularly high masculine ranks; in Czech the \textit{men are childish} stereotype is always ranked as more feminine than masculine; and in Slovenian the \textit{women are weak} stereotype is firmly feminine-coded across models. Other language-specific results vary by model; for instance, \textit{men are strong} has an unusually low masculine rank in Slovak only in EuroLLM, and \textit{women are emotional} is the most feminine stereotype in Polish only in Salamandra.

% For weakness, it may be that the male coding of this stereotype is connected with the prevalence of the `men are strong' stereotype, as terms related to weakness are commonly used in discussions about physical strength.

\subsection{Impact of model size and instruction finetuning on gender stereotyping}
To address our second research question, we first explore how a model's size impacts the degree of stereotyping it exhibits. To isolate the impact of model size alone on stereotyping, we first compare five different sizes of Qwen 2.5 models ranging from 0.5 to 14 billion parameters.\footnote{\href{https://huggingface.co/Qwen/Qwen2.5-0.5B}{\texttt{Qwen 2.5-0.5B}}, \href{https://huggingface.co/Qwen/Qwen2.5-2.5B}{\texttt{Qwen 2.5-1.5B}}, 
\href{https://huggingface.co/Qwen/Qwen2.5-3B}{\texttt{Qwen 2.5-3B}}, \href{https://huggingface.co/Qwen/Qwen2.5-7B}{\texttt{Qwen 2.5-7B}} and \href{https://huggingface.co/Qwen/Qwen2.5-14B}{\texttt{Qwen 2.5-14B}}. We also test the instruct variants of each model.} For each model and each of the 30 EuroGEST languages, we first calculate $q_i$ for each stereotype $i$ and then calculate the \textbf{inclination} (see \Cref{sec:method}) of $q_i$ towards the stereotypical gender of $i$. We average these scores across all languages to obtain an overall measure of the strength of each individual stereotype.

\begin{figure}[h]
    \centering
    \includegraphics[width=1\linewidth]{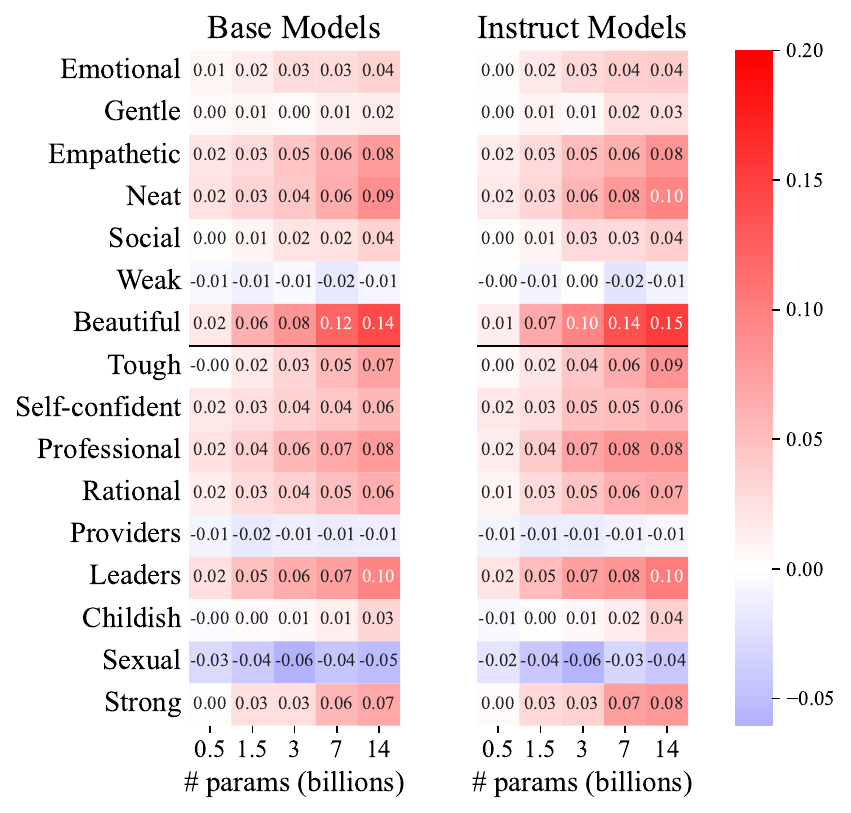}
    \caption{Divergence of $q_i$ scores for each stereotype from proxy default masculine rate towards stereotypical gender for feminine (top) and masculine (bottom) stereotypes in five sizes of Qwen 2.5 models.}
    \label{fig:qwen_size}
\end{figure}

We report results for both base and instruct models in \Cref{fig:qwen_size}. The Qwen 2.5 models clearly reproduce the same strongest and weakest gender stereotypes as the three European-centric models in \Cref{fig:heatmaps}. There is also a visible increase in stereotype strength with model size, as all stereotypes apart from the three weakly encoded ones (\textit{women are weak, men are providers and men are sexual}) show greater inclination towards stereotypical sentences as model size increases.

Finally, we compute $g_s$ scores for all languages for the 13 models already evaluated and for four smaller EuroLLM and Salamandra models,\footnote{\href{https://huggingface.co/utter-project/EuroLLM-1.7B}{\texttt{EuroLLM 1.7B}},  \href{https://huggingface.co/utter-project/EuroLLM-1.7B-Instruct}{\texttt{EuroLLM 1.7B-Instruct}}, \href{https://huggingface.co/blog/eurollm-team/eurollm-9b}{\texttt{EuroLLM 9B}}, \href{https://huggingface.co/BSC-LT/salamandra-2b-instruct}{\texttt{Salamandra 2B-Instruct}}}, six Llama models\footnote{\href{https://huggingface.co/meta-llama/Llama-3.2-1B}{\texttt{Llama 3.1 1B}},  \href{https://huggingface.co/meta-llama/Llama-3.2-3B}{\texttt{Llama 3.1 3B}} and  \href{https://huggingface.co/meta-llama/Meta-Llama-3-8B}{\texttt{Llama 3.1 8B}}. We also test the instruct variants of these models.} and Aya Expanse.\footnote{\href{https://huggingface.co/CohereLabs/aya-expanse-8b}{\texttt{Aya Expanse 8B Instruct}}} We select these models in order to evaluate a diverse set of multilingual model families and sizes, within the scope of our available resources for inference.

\begin{figure}[ht]
    \centering
    \includegraphics[width=1\linewidth]{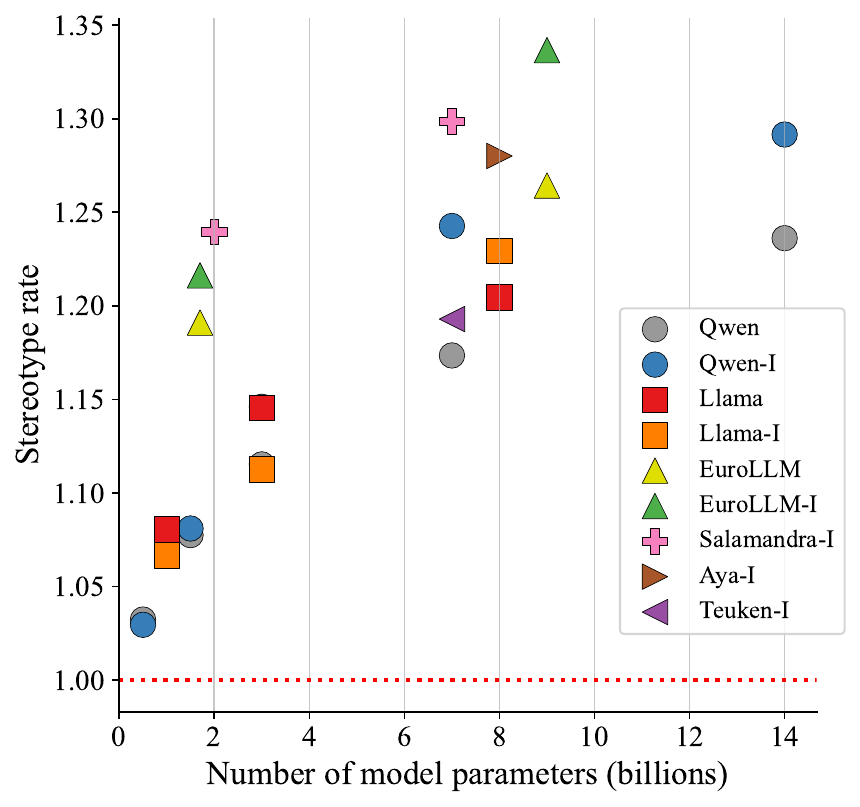}
    \caption{Average stereotype rates of base and instruct models across all languages in EuroGEST. $g_s$ of 1.0 (dotted red line) is indicative of no stereotyping.}
    \label{fig:gs_averages}
\end{figure}

We show the average $g_s$ scores over all languages in \Cref{fig:gs_averages} (with $g_s$ scores per language shown in \Cref{sec:app_5}). Even the smallest models demonstrate stereotypical reasoning, but $g_s$ increases consistently with model size across all families. Where both base and instruct models are available, instruction-finetuning does not appear to have uniformly decreased gender bias, and in some the instruct models exhibit more stereotypical reasoning than their base model counterparts. We also note that models with broader coverage of EuroGEST languages (such as EuroLLM, Salamandra, Teuken, and Aya) tend to exhibit higher average $g_s$ scores than Qwen and Llama. These high $g_s$ scores likely reflect the fact that these models are more highly performant in general on the full range of European languages, compared to the commercial models. This hypothesis is supported by the per-language $g_s$ scores (\Cref{fig:all_langs_g_s_scores}), which show that all models generally have higher $g_s$ scores on languages which they formally support, compared to those for which they only have latent abilities.

\section{Discussion} 
Our dataset creation process highlights both the promise and the challenges of scaling gender bias evaluation tools across a wide range of languages. It also raises a central methodological question: what constitutes data that is ``good enough'' for evaluating LLMs for bias? Even in the relatively well-resourced languages represented in EuroGEST, the complexities of morphological gender marking complicate the process of benchmark data translation. However, with the exception of Maltese, our synthetic data generation method was positively evaluated by professional translators in each language, and produces data that we think is sufficiently robust to illustrate systematic gender biases in LLMs across a broader range of languages than has previously been explored. Even with the Maltese dataset of limited quality, we still see evidence of gender stereotyping by the LLMs we evaluate in Maltese (Figures \ref{fig:heatmaps} and \ref{fig:all_langs_g_s_scores}).

Our multilingual bias evaluations also contend with the difficulties of comparing token likelihoods across models and languages, given the fundamentally different distributions of gendered terms and the varied ways in which gender is expressed morphologically, and differences in tokenisation which may impact raw likelihoods of certain terms. By building on the methods developed in \citet{pikuliak_women_2024} to handle these difficulties, we convincingly show that 13 out of the 16 gendered stereotypes we investigate are consistently present in the internal representations of multilingual LLMs, across all 24 models and 30 languages studied. Portrayals of women as \textit{beautiful}, \textit{empathetic}, and \textit{neat}, and portrayals of men as \textit{leaders}, \textit{strong}, \textit{professional}, and \textit{tough} emerge as the most strongly encoded stereotypes across all models and languages (\Cref{fig:per_stereotype_stats}). These findings align with those of \citet{pikuliak_women_2024}, who observed similar patterns across masked, generative, and translation models. We also replicate their observation that \textit{men are sexual} stereotype sentences are more commonly associated with the feminine grammatical gender (Figures \ref{fig:per_stereotype_stats} and \ref{fig:qwen_size}), likely reflecting the broader sexualisation of women in text \citep{pikuliak_women_2024}.

We further find that larger models generally exhibit stronger gender stereotyping (Figures \ref{fig:qwen_size} and \ref{fig:gs_averages}), consistent with prior work \citep{pikuliak_women_2024, tal-etal-2022-fewer}. This is particularly true in languages where their overall performance is strong. This is intuitive: gender biases emerge as complex distributional patterns in linguistic data, and more powerful models are better equipped to capture them. Models with more multilingual training data will have been exposed to a greater range of multilingual and multicultural expressions of gender stereotypes in their training data. Larger models are also better at modelling morphological and semantic patterns across languages, and we can think of gender  stereotyping as complex patterns which can be inducted by LLMs from training data. Importantly, we also observe that instruction-finetuned models often display stronger gender stereotyping than their base-model counterparts. This underscores the unpredictable effects of instruction tuning, which may inadvertently exacerbate harmful representational patterns in some languages, even as it mitigates them in others.

The persistence and salience of these stereotypes in multilingual LLMs may contribute to a range of representational harms as they are deployed in practice, including erasing the visibility of men and women in different roles and contexts and reinforcing discriminatory behaviour and assumptions over time. The subtlety of these biases and the different ways in which they are expressed across languages makes it difficult to evaluate them in practical downstream tasks, yet the need for robust, multilingual evaluation and mitigation grows more urgent as models increase in size and capability. We hope that EuroGEST will provide a foundation for research into how training data, modelling choices and alignment strategies impacts gender bias in multilingual LLMs by offering a resource that enables the systematic evaluation of gender stereotypes in LLMs across languages. Ultimately, consistent cross-lingual mitigation and evaluation strategies will be essential to ensure that increasingly powerful LLMs do not entrench or amplify gendered harms.

\section{Conclusion}

As LLMs become more powerful and multilingual, it is increasingly important to devise robust evaluation methods to understand how they encode complex social constructs across languages and to minimise the risks of bias and discrimination. With EuroGEST, we extend and release an existing gender bias benchmark dataset \citep{pikuliak_women_2024} in 30 European languages. We also document our resource-efficient method for rapidly and sensitively scaling benchmark data across multiple languages. Beyond gender, this approach may also benefit other areas of responsible AI where language coverage remains a critical gap.

Using EuroGEST, we demonstrate that six families of LLMs systematically encode at least 13 gendered stereotypes. We also show that larger and more powerful models exhibit stronger stereotypical biases and reasoning on supported languages and that instruction-finetuned models are sometimes more biased than their base counterparts. These findings highlight the need for mitigation strategies that are both cross-lingual and sensitive to the diversity of gender representations.

This work fills an urgent gap in the lack of multilingual bias evaluation resources. However, its linguistic breadth necessarily limits its depth, and our one-size-fits-all approach certainly cannot capture the full diversity of how gender bias is expressed and experienced by LLM users across all the languages we consider. We hope that our findings will motivate sustained, participatory efforts with gender minorities and experts across diverse linguistic and cultural contexts to develop evaluation methods and resources that move beyond surface-level benchmarking towards more inclusive and socially grounded approaches.

\section*{Acknowledgements}
This work was supported by the UKRI AI Centre for Doctoral Training in Designing Responsible Natural Language Processing, grant number EP/Y030656/1; EU Horizon Europe Research and Innovation programme grant No 101070631 and UKRI under the UK HE funding grant No 10039436; and the National Science Centre, Poland 2023/49/N/ST6/02691. For the purpose of open access, the authors have applied a Creative Commons Attribution (CC BY) licence to any Author Accepted Manuscript version arising. 

The authors would like to sincerely thank Alessandra Terranova and Lara Dal Molin (University of Edinburgh) for their participation in a pilot version of the validation study and for providing valuable feedback. We are also grateful to Ben Peters (Instituto de Telecomunicações/INESC-ID) for his careful review of EuroGEST’s Portuguese samples, which provided insightful observations on potential limitations and systematic patterns in the gender heuristics employed.

\section*{Limitations}
\label{sec:limitations}

% 

% Interestingly, nearly half of the 120 EuroGEST sentences which translators reported as incorrectly labelled for gender were masculine sentences which had been mislabelled as neutral by our automated system. This is likely because the forced masculine and feminine translations we produced with Google API defaulted to the masculine variant of the sentence even for the feminine-prompted sentence, producing two identical sentences which therefore appeared to be gender-neutral. 

% Multilingual gender bias methods must also tackle the fact that many languages with grammatical gender lack widely-accepted inflection conventions for people of diverse gender groups \citep{saunders_reducing_2020, ackerman_syntactic_2019}, meaning that gender bias benchmarks designed specifically to be gender-inclusive (e.g. \citet{munro_detecting_2020}) can be difficult to scale multilingually.   

% Some work explores biases against non-binary individuals by measuring biases against English third-person singular pronouns “they” and “theirs” in co-reference tasks  However, as highlighted by , many gendered languages lack widely-accepted non-binary inflection conventions, making it difficult to integrate more diverse gender groups into empirical methods that rely upon morphological inflection to detect gendered categories. 

\paragraph{Scope of biases examined in this work}
We investigate sixteen specific gendered stereotypes, originally identified in previous work by gender studies experts and literature reviews, by comparing the likelihood of stereotypical and anti-stereotypical sentences during text generation. These are but a small subset of the ways gender bias may arise as LLMs are applied to specific tasks or contexts; future work could further examine how these biases connect with concrete gendered harms experienced by users in practice \citep{zhou2023publicperceptionsgenderbias, williams2024stereo}, particularly as LLMs are deployed across different languages and sociocultural contexts. 

We investigate stereotypes commonly held about men and women, but we do not address LLM biases about people of other genders, as explored in previous work \citep{blodgett_language_2020, dev_harms_2021, goldfarb-tarrant_this_2023, talat_you_2022, munro_detecting_2020}. This decision is primarily motivated by the lack of standardised gender-inclusive inflection conventions in the gendered languages in our set of target languages. However, we acknowledge that this practical decision risks reinforcing an exclusive or binary understanding of gender, overlooking or minimising ways in which LLM biases may impact nonbinary and gender-diverse individuals. To address this, we seek to make clear that EuroGEST measures only specific gendered stereotypes about men and women, not `gender bias' in its entirety. We also use the gender-inclusive terms `masculine' and `feminine' throughout the work, rather than `male' and `female'. We hope in future work to expand our method further to include more diverse gender categories, and have already begun to consult language experts for appropriate constructions in each of EuroGEST's languages for this next stage.

Finally, we do not incorporate any intersectional analysis, but acknowledge that many other social demographic factors intersect with and in some cases exacerbate gender biases in LLMs. Neglecting intersectionality may obscure compounded or unique forms of bias encoded in LLMs, particularly in multilingual contexts. Scaling gender-diverse and intersectional analyses in multilingual gender bias detection is an important direction for future work, and will provide a more holistic picture of LLMs' social biases.

\paragraph{English-centricity}

While European countries share many societal and economic similarities, the stereotypes we examine may reflect norms more aligned with Anglophone contexts. There is a risk that EuroGEST underrepresents culturally specific stereotypes prevalent in different European regions, potentially overlooking how LLMs replicate localised biases. Moreover, many languages in EuroGEST are spoken in non-European countries where gender norms may differ substantially. Applying EuroGEST in such contexts risks drawing misleading conclusions about model behaviour across global populations.

A further limitation is the reliance on English-centric noun- and pronoun-templates -- such as ‘S’, he said and ‘S’, the woman said -- which may be less grammatical or tokenised in awkward or inconsistent ways in some languages. There is a risk that unnatural tokenization or grammatical mismatches could affect the accuracy and fairness of bias measurements obtained by using EuroGEST. In future work, language-specific templates that better reflect organic usage and control for tokenisation should be developed.

\paragraph{Automatic translation}

We utilise automatic translation for resource-efficient scaling of EuroGEST, and while we employ quality evaluation through both COMET-QE filtering and human validation of a subset of the dataset, we cannot guarantee that all EuroGEST sentences are correctly and fluently translated into each language. Automatic translation is not as effective for the lower-resourced languages in the dataset, which reduces both the quality of the translations (\Cref{fig:val_stats}) and the number of available sentences for evaluating models in these languages (\Cref{fig:per_stereotype_stats}).

To identify gendered minimal pairs, we rely on a one-size-fits-all heuristic of less than two letters different on less than one word. This results in both under-inclusion of legitimate gendered pairs where gender is expressed on multiple words or through longer suffixes than allowed by the heuristic, but also over-inclusion of illegitimate pairs. For example, in some cases a random error or variation in translation results in different words which happen to differ by only two letters, even though they are not gendered pairs. Furthermore, for sentences referring to romantic relationships in some languages, this heuristic also captures cases where the object of the sentence (rather than the subject) is gendered in ways which reflect an assumption of heterosexuality by the Google Translate API.\footnote{For example, the object \textit{`artist'} of the English sentence \textit{``I asked the artist on a date''} is usually translated into languages like Portuguese with feminine gender \textit{`artista'} when the subject is indicated to be masculine, and masculine gender \textit{`artisto'} when the subject is indicated to be feminine, reflecting an assumption of heterosexuality.} While this is certainly an issue we would like to address in future work, it is unfortunately quite likely that the LLMs we are measuring \textit{also} make the same heterosexual assumptions (e.g. interpreting the sentence with feminine object \textit{`artista'} to be indicative of a masculine subject, and masculine object \textit{`artisto'} to be indicative of a feminine subject). The existence of these samples therefore does not necessarily undermine the reliability of the data and the results, depending on how the sentences are used in practice.

Future work could develop language-specific heuristics which more carefully avoid these error cases and retain a higher proportion of the legitimate gendered minimal pairs in each language, depending on the number of words usually gendered in a given sentence or the length of gendered suffixes in that language. 

% These differences reflect that the models have been trained on different data and subjected to different finetuning strategies, and also that the distribution of masculine and feminine variants of gendered terms in text varies across languages. The ways in which the masculine and feminine variants of each sentence are expressed and tokenised also varies across languages and models, impacting the relative probabilities of different gendered variants. For example, the French masculine noun template (\textit{dit l’homme}) is more complex than the feminine one (d\textit{it la femme}) in its inclusion of an elided article and apostrophe, possibly lowering the relative probability of the masculine template compared to the feminine template overall. Despite these differences, the gap between masculine rates for masculine vs. feminine stereotypes remains a reliable indicator of encoded bias, and this is where we focus the rest of our analysis. 

\bibliography{eurogest_paper}

@article{mastromichalakis2025assumedidentitiesquantifyinggender,
      title={Assumed Identities: Quantifying Gender Bias in Machine Translation of Gender-Ambiguous Occupational Terms}, 
      author={Orfeas Menis Mastromichalakis and Giorgos Filandrianos and Maria Symeonaki and Giorgos Stamou},
      year={2025},
      journal={arXiv preprint arXiv:2503.04372},
      archivePrefix={arXiv},
      primaryClass={cs.CL},
      url={https://arxiv.org/abs/2503.04372}, 
}

@article{martins2025eurollm9btechnicalreport,
      title={EuroLLM-9B: Technical Report}, 
      author={Pedro Henrique Martins and João Alves and Patrick Fernandes and Nuno M. Guerreiro and Ricardo Rei and Amin Farajian and Mateusz Klimaszewski and Duarte M. Alves and José Pombal and Nicolas Boizard and Manuel Faysse and Pierre Colombo and François Yvon and Barry Haddow and José G. C. de Souza and Alexandra Birch and André F. T. Martins},
      year={2025},
      journal={arXiv preprint arXiv:2506.04079},
      archivePrefix={arXiv},
      primaryClass={cs.CL},
      url={https://arxiv.org/abs/2506.04079}, 
}

@article{gonzalezagirre2025salamandratechnicalreport,
      title={Salamandra Technical Report}, 
      author={Aitor Gonzalez-Agirre and Marc Pàmies and Joan Llop and Irene Baucells and Severino Da Dalt and Daniel Tamayo and José Javier Saiz and Ferran Espuña and Jaume Prats and Javier Aula-Blasco and Mario Mina and Iñigo Pikabea and Adrián Rubio and Alexander Shvets and Anna Sallés and Iñaki Lacunza and Jorge Palomar and Júlia Falcão and Lucía Tormo and Luis Vasquez-Reina and Montserrat Marimon and Oriol Pareras and Valle Ruiz-Fernández and Marta Villegas},
      year={2025},
      journal={arXiv preprint arXiv:2502.08489},
      archivePrefix={arXiv},
      primaryClass={cs.CL},
      url={https://arxiv.org/abs/2502.08489}, 
}

@article{ali2025teuken7bbaseteuken7binstructeuropean,
      title={Teuken-7B-Base \& Teuken-7B-Instruct: Towards European LLMs}, 
      author={Mehdi Ali and Michael Fromm and Klaudia Thellmann and Jan Ebert and Alexander Arno Weber and Richard Rutmann and Charvi Jain and Max Lübbering and Daniel Steinigen and Johannes Leveling and Katrin Klug and Jasper Schulze Buschhoff and Lena Jurkschat and Hammam Abdelwahab and Benny Jörg Stein and Karl-Heinz Sylla and Pavel Denisov and Nicolo' Brandizzi and Qasid Saleem and Anirban Bhowmick and Lennard Helmer and Chelsea John and Pedro Ortiz Suarez and Malte Ostendorff and Alex Jude and Lalith Manjunath and Samuel Weinbach and Carolin Penke and Oleg Filatov and Fabio Barth and Paramita Mirza and Lucas Weber and Ines Wendler and Rafet Sifa and Fabian Küch and Andreas Herten and René Jäkel and Georg Rehm and Stefan Kesselheim and Joachim Köhler and Nicolas Flores-Herr},
      year={2025},
      journal={arXiv preprint arXiv:2410.03730},
      archivePrefix={arXiv},
      primaryClass={cs.CL},
      url={https://arxiv.org/abs/2410.03730}, 
}

@inproceedings{tal-etal-2022-fewer,
    title = "Fewer Errors, but More Stereotypes? The Effect of Model Size on Gender Bias",
    author = "Tal, Yarden  and
      Magar, Inbal  and
      Schwartz, Roy",
    editor = "Hardmeier, Christian  and
      Basta, Christine  and
      Costa-juss{\`a}, Marta R.  and
      Stanovsky, Gabriel  and
      Gonen, Hila",
    booktitle = "Proceedings of the 4th Workshop on Gender Bias in Natural Language Processing (GeBNLP)",
    month = jul,
    year = "2022",
    address = "Seattle, Washington",
    publisher = "Association for Computational Linguistics",
    url = "https://aclanthology.org/2022.gebnlp-1.13/",
    doi = "10.18653/v1/2022.gebnlp-1.13",
    pages = "112--120"
}

@article{qwen2025qwen25technicalreport,
      title={Qwen2.5 Technical Report}, 
      author={An Yang and Baosong Yang and Beichen Zhang and Binyuan Hui and Bo Zheng and Bowen Yu and Chengyuan Li and Dayiheng Liu and Fei Huang and Haoran Wei and Huan Lin and Jian Yang and Jianhong Tu and Jianwei Zhang and Jianxin Yang and Jiaxi Yang and Jingren Zhou and Junyang Lin and Kai Dang and Keming Lu and Keqin Bao and Kexin Yang and Le Yu and Mei Li and Mingfeng Xue and Pei Zhang and Qin Zhu and Rui Men and Runji Lin and Tianhao Li and Tianyi Tang and Tingyu Xia and Xingzhang Ren and Xuancheng Ren and Yang Fan and Yang Su and Yichang Zhang and Yu Wan and Yuqiong Liu and Zeyu Cui and Zhenru Zhang and Zihan Qiu},
    journal={arXiv preprint arXiv:2412.15115},
      year={2025},
      eprint={2412.15115},
      archivePrefix={arXiv},
      primaryClass={cs.CL},
      url={https://arxiv.org/abs/2412.15115}, 
}

@article{dang2024ayaexpansecombiningresearch,
      title={Aya Expanse: Combining Research Breakthroughs for a New Multilingual Frontier}, 
      author={John Dang and Shivalika Singh and Daniel D'souza and Arash Ahmadian and Alejandro Salamanca and Madeline Smith and Aidan Peppin and Sungjin Hong and Manoj Govindassamy and Terrence Zhao and Sandra Kublik and Meor Amer and Viraat Aryabumi and Jon Ander Campos and Yi-Chern Tan and Tom Kocmi and Florian Strub and Nathan Grinsztajn and Yannis Flet-Berliac and Acyr Locatelli and Hangyu Lin and Dwarak Talupuru and Bharat Venkitesh and David Cairuz and Bowen Yang and Tim Chung and Wei-Yin Ko and Sylvie Shang Shi and Amir Shukayev and Sammie Bae and Aleksandra Piktus and Roman Castagné and Felipe Cruz-Salinas and Eddie Kim and Lucas Crawhall-Stein and Adrien Morisot and Sudip Roy and Phil Blunsom and Ivan Zhang and Aidan Gomez and Nick Frosst and Marzieh Fadaee and Beyza Ermis and Ahmet Üstün and Sara Hooker},
      year={2024},
      journal={arXiv preprint arXiv:2412.04261},
      archivePrefix={arXiv},
      primaryClass={cs.CL},
      url={https://arxiv.org/abs/2412.04261}, 
}

@article{grattafiori2024llama3herdmodels,
      title={The Llama 3 Herd of Models}, 
      author={Aaron Grattafiori and Abhimanyu Dubey and Abhinav Jauhri and Abhinav Pandey and Abhishek Kadian and Ahmad Al-Dahle and Aiesha Letman and Akhil Mathur and Alan Schelten and Alex Vaughan and Amy Yang and Angela Fan and Anirudh Goyal and Anthony Hartshorn and Aobo Yang and Archi Mitra and Archie Sravankumar and Artem Korenev and Arthur Hinsvark and Arun Rao and Aston Zhang and Aurelien Rodriguez and Austen Gregerson and Ava Spataru and Baptiste Roziere and Bethany Biron and Binh Tang and Bobbie Chern and Charlotte Caucheteux and Chaya Nayak and Chloe Bi and Chris Marra and Chris McConnell and Christian Keller and Christophe Touret and Chunyang Wu and Corinne Wong and Cristian Canton Ferrer and Cyrus Nikolaidis and Damien Allonsius and Daniel Song and Danielle Pintz and Danny Livshits and Danny Wyatt and David Esiobu and Dhruv Choudhary and Dhruv Mahajan and Diego Garcia-Olano and Diego Perino and Dieuwke Hupkes and Egor Lakomkin and Ehab AlBadawy and Elina Lobanova and Emily Dinan and Eric Michael Smith and Filip Radenovic and Francisco Guzmán and Frank Zhang and Gabriel Synnaeve and Gabrielle Lee and Georgia Lewis Anderson and Govind Thattai and Graeme Nail and Gregoire Mialon and Guan Pang and Guillem Cucurell and Hailey Nguyen and Hannah Korevaar and Hu Xu and Hugo Touvron and Iliyan Zarov and Imanol Arrieta Ibarra and Isabel Kloumann and Ishan Misra and Ivan Evtimov and Jack Zhang and Jade Copet and Jaewon Lee and Jan Geffert and Jana Vranes and Jason Park and Jay Mahadeokar and Jeet Shah and Jelmer van der Linde and Jennifer Billock and Jenny Hong and Jenya Lee and Jeremy Fu and Jianfeng Chi and Jianyu Huang and Jiawen Liu and Jie Wang and Jiecao Yu and Joanna Bitton and Joe Spisak and Jongsoo Park and Joseph Rocca and Joshua Johnstun and Joshua Saxe and Junteng Jia and Kalyan Vasuden Alwala and Karthik Prasad and Kartikeya Upasani and Kate Plawiak and Ke Li and Kenneth Heafield and Kevin Stone and Khalid El-Arini and Krithika Iyer and Kshitiz Malik and Kuenley Chiu and Kunal Bhalla and Kushal Lakhotia and Lauren Rantala-Yeary and Laurens van der Maaten and Lawrence Chen and Liang Tan and Liz Jenkins and Louis Martin and Lovish Madaan and Lubo Malo and Lukas Blecher and Lukas Landzaat and Luke de Oliveira and Madeline Muzzi and Mahesh Pasupuleti and Mannat Singh and Manohar Paluri and Marcin Kardas and Maria Tsimpoukelli and Mathew Oldham and Mathieu Rita and Maya Pavlova and Melanie Kambadur and Mike Lewis and Min Si and Mitesh Kumar Singh and Mona Hassan and Naman Goyal and Narjes Torabi and Nikolay Bashlykov and Nikolay Bogoychev and Niladri Chatterji and Ning Zhang and Olivier Duchenne and Onur Çelebi and Patrick Alrassy and Pengchuan Zhang and Pengwei Li and Petar Vasic and Peter Weng and Prajjwal Bhargava and Pratik Dubal and Praveen Krishnan and Punit Singh Koura and Puxin Xu and Qing He and Qingxiao Dong and Ragavan Srinivasan and Raj Ganapathy and Ramon Calderer and Ricardo Silveira Cabral and Robert Stojnic and Roberta Raileanu and Rohan Maheswari and Rohit Girdhar and Rohit Patel and Romain Sauvestre and Ronnie Polidoro and Roshan Sumbaly and Ross Taylor and Ruan Silva and Rui Hou and Rui Wang and Saghar Hosseini and Sahana Chennabasappa and Sanjay Singh and Sean Bell and Seohyun Sonia Kim and Sergey Edunov and Shaoliang Nie and Sharan Narang and Sharath Raparthy and Sheng Shen and Shengye Wan and Shruti Bhosale and Shun Zhang and Simon Vandenhende and Soumya Batra and Spencer Whitman and Sten Sootla and Stephane Collot and Suchin Gururangan and Sydney Borodinsky and Tamar Herman and Tara Fowler and Tarek Sheasha and Thomas Georgiou and Thomas Scialom and Tobias Speckbacher and Todor Mihaylov and Tong Xiao and Ujjwal Karn and Vedanuj Goswami and Vibhor Gupta and Vignesh Ramanathan and Viktor Kerkez and Vincent Gonguet and Virginie Do and Vish Vogeti and Vítor Albiero and Vladan Petrovic and Weiwei Chu and Wenhan Xiong and Wenyin Fu and Whitney Meers and Xavier Martinet and Xiaodong Wang and Xiaofang Wang and Xiaoqing Ellen Tan and Xide Xia and Xinfeng Xie and Xuchao Jia and Xuewei Wang and Yaelle Goldschlag and Yashesh Gaur and Yasmine Babaei and Yi Wen and Yiwen Song and Yuchen Zhang and Yue Li and Yuning Mao and Zacharie Delpierre Coudert and Zheng Yan and Zhengxing Chen and Zoe Papakipos and Aaditya Singh and Aayushi Srivastava and Abha Jain and Adam Kelsey and Adam Shajnfeld and Adithya Gangidi and Adolfo Victoria and Ahuva Goldstand and Ajay Menon and Ajay Sharma and Alex Boesenberg and Alexei Baevski and Allie Feinstein and Amanda Kallet and Amit Sangani and Amos Teo and Anam Yunus and Andrei Lupu and Andres Alvarado and Andrew Caples and Andrew Gu and Andrew Ho and Andrew Poulton and Andrew Ryan and Ankit Ramchandani and Annie Dong and Annie Franco and Anuj Goyal and Aparajita Saraf and Arkabandhu Chowdhury and Ashley Gabriel and Ashwin Bharambe and Assaf Eisenman and Azadeh Yazdan and Beau James and Ben Maurer and Benjamin Leonhardi and Bernie Huang and Beth Loyd and Beto De Paola and Bhargavi Paranjape and Bing Liu and Bo Wu and Boyu Ni and Braden Hancock and Bram Wasti and Brandon Spence and Brani Stojkovic and Brian Gamido and Britt Montalvo and Carl Parker and Carly Burton and Catalina Mejia and Ce Liu and Changhan Wang and Changkyu Kim and Chao Zhou and Chester Hu and Ching-Hsiang Chu and Chris Cai and Chris Tindal and Christoph Feichtenhofer and Cynthia Gao and Damon Civin and Dana Beaty and Daniel Kreymer and Daniel Li and David Adkins and David Xu and Davide Testuggine and Delia David and Devi Parikh and Diana Liskovich and Didem Foss and Dingkang Wang and Duc Le and Dustin Holland and Edward Dowling and Eissa Jamil and Elaine Montgomery and Eleonora Presani and Emily Hahn and Emily Wood and Eric-Tuan Le and Erik Brinkman and Esteban Arcaute and Evan Dunbar and Evan Smothers and Fei Sun and Felix Kreuk and Feng Tian and Filippos Kokkinos and Firat Ozgenel and Francesco Caggioni and Frank Kanayet and Frank Seide and Gabriela Medina Florez and Gabriella Schwarz and Gada Badeer and Georgia Swee and Gil Halpern and Grant Herman and Grigory Sizov and Guangyi and Zhang and Guna Lakshminarayanan and Hakan Inan and Hamid Shojanazeri and Han Zou and Hannah Wang and Hanwen Zha and Haroun Habeeb and Harrison Rudolph and Helen Suk and Henry Aspegren and Hunter Goldman and Hongyuan Zhan and Ibrahim Damlaj and Igor Molybog and Igor Tufanov and Ilias Leontiadis and Irina-Elena Veliche and Itai Gat and Jake Weissman and James Geboski and James Kohli and Janice Lam and Japhet Asher and Jean-Baptiste Gaya and Jeff Marcus and Jeff Tang and Jennifer Chan and Jenny Zhen and Jeremy Reizenstein and Jeremy Teboul and Jessica Zhong and Jian Jin and Jingyi Yang and Joe Cummings and Jon Carvill and Jon Shepard and Jonathan McPhie and Jonathan Torres and Josh Ginsburg and Junjie Wang and Kai Wu and Kam Hou U and Karan Saxena and Kartikay Khandelwal and Katayoun Zand and Kathy Matosich and Kaushik Veeraraghavan and Kelly Michelena and Keqian Li and Kiran Jagadeesh and Kun Huang and Kunal Chawla and Kyle Huang and Lailin Chen and Lakshya Garg and Lavender A and Leandro Silva and Lee Bell and Lei Zhang and Liangpeng Guo and Licheng Yu and Liron Moshkovich and Luca Wehrstedt and Madian Khabsa and Manav Avalani and Manish Bhatt and Martynas Mankus and Matan Hasson and Matthew Lennie and Matthias Reso and Maxim Groshev and Maxim Naumov and Maya Lathi and Meghan Keneally and Miao Liu and Michael L. Seltzer and Michal Valko and Michelle Restrepo and Mihir Patel and Mik Vyatskov and Mikayel Samvelyan and Mike Clark and Mike Macey and Mike Wang and Miquel Jubert Hermoso and Mo Metanat and Mohammad Rastegari and Munish Bansal and Nandhini Santhanam and Natascha Parks and Natasha White and Navyata Bawa and Nayan Singhal and Nick Egebo and Nicolas Usunier and Nikhil Mehta and Nikolay Pavlovich Laptev and Ning Dong and Norman Cheng and Oleg Chernoguz and Olivia Hart and Omkar Salpekar and Ozlem Kalinli and Parkin Kent and Parth Parekh and Paul Saab and Pavan Balaji and Pedro Rittner and Philip Bontrager and Pierre Roux and Piotr Dollar and Polina Zvyagina and Prashant Ratanchandani and Pritish Yuvraj and Qian Liang and Rachad Alao and Rachel Rodriguez and Rafi Ayub and Raghotham Murthy and Raghu Nayani and Rahul Mitra and Rangaprabhu Parthasarathy and Raymond Li and Rebekkah Hogan and Robin Battey and Rocky Wang and Russ Howes and Ruty Rinott and Sachin Mehta and Sachin Siby and Sai Jayesh Bondu and Samyak Datta and Sara Chugh and Sara Hunt and Sargun Dhillon and Sasha Sidorov and Satadru Pan and Saurabh Mahajan and Saurabh Verma and Seiji Yamamoto and Sharadh Ramaswamy and Shaun Lindsay and Shaun Lindsay and Sheng Feng and Shenghao Lin and Shengxin Cindy Zha and Shishir Patil and Shiva Shankar and Shuqiang Zhang and Shuqiang Zhang and Sinong Wang and Sneha Agarwal and Soji Sajuyigbe and Soumith Chintala and Stephanie Max and Stephen Chen and Steve Kehoe and Steve Satterfield and Sudarshan Govindaprasad and Sumit Gupta and Summer Deng and Sungmin Cho and Sunny Virk and Suraj Subramanian and Sy Choudhury and Sydney Goldman and Tal Remez and Tamar Glaser and Tamara Best and Thilo Koehler and Thomas Robinson and Tianhe Li and Tianjun Zhang and Tim Matthews and Timothy Chou and Tzook Shaked and Varun Vontimitta and Victoria Ajayi and Victoria Montanez and Vijai Mohan and Vinay Satish Kumar and Vishal Mangla and Vlad Ionescu and Vlad Poenaru and Vlad Tiberiu Mihailescu and Vladimir Ivanov and Wei Li and Wenchen Wang and Wenwen Jiang and Wes Bouaziz and Will Constable and Xiaocheng Tang and Xiaojian Wu and Xiaolan Wang and Xilun Wu and Xinbo Gao and Yaniv Kleinman and Yanjun Chen and Ye Hu and Ye Jia and Ye Qi and Yenda Li and Yilin Zhang and Ying Zhang and Yossi Adi and Youngjin Nam and Yu and Wang and Yu Zhao and Yuchen Hao and Yundi Qian and Yunlu Li and Yuzi He and Zach Rait and Zachary DeVito and Zef Rosnbrick and Zhaoduo Wen and Zhenyu Yang and Zhiwei Zhao and Zhiyu Ma},
      year={2024},
      journal={arXiv preprint arXiv:2407.21783},
      archivePrefix={arXiv},
      primaryClass={cs.AI},
      url={https://arxiv.org/abs/2407.21783}, 
}

@inproceedings{rei-etal-2022-cometkiwi,
    title = "{C}omet{K}iwi: {IST}-Unbabel 2022 Submission for the Quality Estimation Shared Task",
    author = "Rei, Ricardo  and
      Treviso, Marcos  and
      Guerreiro, Nuno M.  and
      Zerva, Chrysoula  and
      Farinha, Ana C  and
      Maroti, Christine  and
      C. de Souza, Jos{\'e} G.  and
      Glushkova, Taisiya  and
      Alves, Duarte  and
      Coheur, Luisa  and
      Lavie, Alon  and
      Martins, Andr{\'e} F. T.",
    editor = {Koehn, Philipp  and
      Barrault, Lo{\"i}c  and
      Bojar, Ond{\v{r}}ej  and
      Bougares, Fethi  and
      Chatterjee, Rajen  and
      Costa-juss{\`a}, Marta R.  and
      Federmann, Christian  and
      Fishel, Mark  and
      Fraser, Alexander  and
      Freitag, Markus  and
      Graham, Yvette  and
      Grundkiewicz, Roman  and
      Guzman, Paco  and
      Haddow, Barry  and
      Huck, Matthias  and
      Jimeno Yepes, Antonio  and
      Kocmi, Tom  and
      Martins, Andr{\'e}  and
      Morishita, Makoto  and
      Monz, Christof  and
      Nagata, Masaaki  and
      Nakazawa, Toshiaki  and
      Negri, Matteo  and
      N{\'e}v{\'e}ol, Aur{\'e}lie  and
      Neves, Mariana  and
      Popel, Martin  and
      Turchi, Marco  and
      Zampieri, Marcos},
    booktitle = "Proceedings of the Seventh Conference on Machine Translation (WMT)",
    month = dec,
    year = "2022",
    address = "Abu Dhabi, United Arab Emirates (Hybrid)",
    publisher = "Association for Computational Linguistics",
    url = "https://aclanthology.org/2022.wmt-1.60/",
    pages = "634--645"
}

@inproceedings{zhu-etal-2024-multilingual,
    title = "Multilingual Machine Translation with Large Language Models: Empirical Results and Analysis",
    author = "Zhu, Wenhao  and
      Liu, Hongyi  and
      Dong, Qingxiu  and
      Xu, Jingjing  and
      Huang, Shujian  and
      Kong, Lingpeng  and
      Chen, Jiajun  and
      Li, Lei",
    editor = "Duh, Kevin  and
      Gomez, Helena  and
      Bethard, Steven",
    booktitle = "Findings of the Association for Computational Linguistics: NAACL 2024",
    month = jun,
    year = "2024",
    address = "Mexico City, Mexico",
    publisher = "Association for Computational Linguistics",
    url = "https://aclanthology.org/2024.findings-naacl.176/",
    doi = "10.18653/v1/2024.findings-naacl.176",
    pages = "2765--2781"
}

@article{goldfarb-tarrant_this_2023,
	title = {This {Prompt} is {Measuring} {\textless}{MASK}{\textgreater}: {Evaluating} {Bias} {Evaluation} in {Language} {Models}},
	shorttitle = {This {Prompt} is {Measuring} {\textless}{MASK}{\textgreater}},
	url = {http://arxiv.org/abs/2305.12757},
	language = {en},
	urldate = {2024-09-24},
	archivePrefix={arXiv},
	author = {Goldfarb-Tarrant, Seraphina and Ungless, Eddie and Balkir, Esma and Blodgett, Su Lin},
	month = may,
	year = {2023},
	journal={arXiv preprint arXiv:2305.12757 },
}

@article{martins2024eurollmmultilinguallanguagemodels,
      title={EuroLLM: Multilingual Language Models for Europe}, 
      author={Pedro Henrique Martins and Patrick Fernandes and João Alves and Nuno M. Guerreiro and Ricardo Rei and Duarte M. Alves and José Pombal and Amin Farajian and Manuel Faysse and Mateusz Klimaszewski and Pierre Colombo and Barry Haddow and José G. C. de Souza and Alexandra Birch and André F. T. Martins},
      year={2024},
      journal={arXiv preprint arXiv:2409.16235},
      archivePrefix={arXiv},
      primaryClass={cs.CL},
      url={https://arxiv.org/abs/2409.16235}, 
}

@article{sanh_multitask_2022,
  title={Multitask prompted training enables zero-shot task generalization},
  author={Sanh, Victor and Webson, Albert and Raffel, Colin and Bach, Stephen H and Sutawika, Lintang and Alyafeai, Zaid and Chaffin, Antoine and Stiegler, Arnaud and Scao, Teven Le and Raja, Arun and others},
  journal={arXiv preprint arXiv:2110.08207},
  year={2021},
url={https://arxiv.org/abs/2110.08207}
}

@article{rottger_safetyprompts_2024,
	title = {{SafetyPrompts}: a {Systematic} {Review} of {Open} {Datasets} for {Evaluating} and {Improving} {Large} {Language} {Model} {Safety}},
	shorttitle = {{SafetyPrompts}},
	url = {http://arxiv.org/abs/2404.05399},
	language = {en},
	urldate = {2024-10-21},
	archivePrefix={arXiv},
	author = {Röttger, Paul and Pernisi, Fabio and Vidgen, Bertie and Hovy, Dirk},
    journal={arXiv preprint arXiv:2404.05399},
	month = apr,
	year = {2024},
}

@article{gupta_calm_2024,
	title = {{CALM} : {A} {Multi}-task {Benchmark} for {Comprehensive} {Assessment} of {Language} {Model} {Bias}},
	shorttitle = {{CALM}},
	url = {http://arxiv.org/abs/2308.12539},
	language = {en},
	urldate = {2024-10-22},
	archivePrefix={arXiv},
	author = {Gupta, Vipul and Venkit, Pranav Narayanan and Laurençon, Hugo and Wilson, Shomir and Passonneau, Rebecca J.},
	month = aug,
	year = {2024},
	journal={arXiv preprint arXiv:2308.12539},
}

@inproceedings{smith_im_2022,
	address = {Abu Dhabi, United Arab Emirates},
	title = {“{I}’m sorry to hear that”: {Finding} {New} {Biases} in {Language} {Models} with a {Holistic} {Descriptor} {Dataset}},
	shorttitle = {“{I}’m sorry to hear that”},
	url = {https://aclanthology.org/2022.emnlp-main.625},
	doi = {10.18653/v1/2022.emnlp-main.625},
	language = {en},
	urldate = {2024-10-22},
	booktitle = {Proceedings of the 2022 {Conference} on {Empirical} {Methods} in {Natural} {Language} {Processing}},
	publisher = {Association for Computational Linguistics},
	author = {Smith, Eric Michael and Hall, Melissa and Kambadur, Melanie and Presani, Eleonora and Williams, Adina},
	year = {2022},
	pages = {9180--9211},
}

@article{parrish_bbq_2022,
	title = {{BBQ}: {A} {Hand}-{Built} {Bias} {Benchmark} for {Question} {Answering}},
	shorttitle = {{BBQ}},
	url = {http://arxiv.org/abs/2110.08193},
	language = {en},
	urldate = {2024-10-22},
	publisher = {arXiv},
	author = {Parrish, Alicia and Chen, Angelica and Nangia, Nikita and Padmakumar, Vishakh and Phang, Jason and Thompson, Jana and Htut, Phu Mon and Bowman, Samuel R.},
	month = mar,
	year = {2022},
	journal={arXiv preprint arXiv:2110.08193},
}

@article{williams2024stereo,
  title={Stereo-Typing: LLM Chatbots’ Appearance of Typing can Increase Belief in a False Stereotype},
  author={Williams-Ceci, Sterling and Zalmanson, Lior and MACY, MICHAEL W and Naaman, Mor},
  year={2024},
  publisher={OSF}
}

@article{zhou2023publicperceptionsgenderbias,
      title={Public Perceptions of Gender Bias in Large Language Models: Cases of ChatGPT and Ernie}, 
      author={Kyrie Zhixuan Zhou and Madelyn Rose Sanfilippo},
      year={2023},
      journal={arXiv preprint arXiv:2309.09120},
      archivePrefix={arXiv},
      primaryClass={cs.AI},
      url={https://arxiv.org/abs/2309.09120}, 
}

@misc{kirk_bias_2021,
	title = {Bias {Out}-of-the-{Box}: {An} {Empirical} {Analysis} of {Intersectional} {Occupational} {Biases} in {Popular} {Generative} {Language} {Models}},
	shorttitle = {Bias {Out}-of-the-{Box}},
	url = {http://arxiv.org/abs/2102.04130},
	language = {en},
	urldate = {2024-10-22},
	publisher = {arXiv},
	author = {Kirk, Hannah and Jun, Yennie and Iqbal, Haider and Benussi, Elias and Volpin, Filippo and Dreyer, Frederic A. and Shtedritski, Aleksandar and Asano, Yuki M.},
	month = oct,
	year = {2021},
	note = {arXiv:2102.04130},
}

@inproceedings{nangia_crows-pairs_2020,
	address = {Online},
	title = {{CrowS}-{Pairs}: {A} {Challenge} {Dataset} for {Measuring} {Social} {Biases} in {Masked} {Language} {Models}},
	shorttitle = {{CrowS}-{Pairs}},
	url = {https://www.aclweb.org/anthology/2020.emnlp-main.154},
	doi = {10.18653/v1/2020.emnlp-main.154},
	language = {en},
	urldate = {2024-10-22},
	booktitle = {Proceedings of the 2020 {Conference} on {Empirical} {Methods} in {Natural} {Language} {Processing} ({EMNLP})},
	publisher = {Association for Computational Linguistics},
	author = {Nangia, Nikita and Vania, Clara and Bhalerao, Rasika and Bowman, Samuel R.},
	year = {2020},
	pages = {1953--1967},
}

@inproceedings{barikeri_redditbias_2021,
	address = {Online},
	title = {{RedditBias}: {A} {Real}-{World} {Resource} for {Bias} {Evaluation} and {Debiasing} of {Conversational} {Language} {Models}},
	shorttitle = {{RedditBias}},
	url = {https://aclanthology.org/2021.acl-long.151},
	doi = {10.18653/v1/2021.acl-long.151},
	language = {en},
	urldate = {2024-10-22},
	booktitle = {Proceedings of the 59th {Annual} {Meeting} of the {Association} for {Computational} {Linguistics} and the 11th {International} {Joint} {Conference} on {Natural} {Language} {Processing} ({Volume} 1: {Long} {Papers})},
	publisher = {Association for Computational Linguistics},
	author = {Barikeri, Soumya and Lauscher, Anne and Vulić, Ivan and Glavaš, Goran},
	year = {2021},
	pages = {1941--1955},
}

@inproceedings{zhao_gender_2018,
	address = {New Orleans, Louisiana},
	title = {Gender {Bias} in {Coreference} {Resolution}: {Evaluation} and {Debiasing} {Methods}},
	shorttitle = {Gender {Bias} in {Coreference} {Resolution}},
	url = {http://aclweb.org/anthology/N18-2003},
	doi = {10.18653/v1/N18-2003},
	language = {en},
	urldate = {2024-10-22},
	booktitle = {Proceedings of the 2018 {Conference} of the {North} {American} {Chapter} of           the {Association} for {Computational} {Linguistics}: {Human} {Language}           {Technologies}, {Volume} 2 ({Short} {Papers})},
	publisher = {Association for Computational Linguistics},
	author = {Zhao, Jieyu and Wang, Tianlu and Yatskar, Mark and Ordonez, Vicente and Chang, Kai-Wei},
	year = {2018},
	pages = {15--20},
}

@inproceedings{rudinger_gender_2018,
	address = {New Orleans, Louisiana},
	title = {Gender {Bias} in {Coreference} {Resolution}},
	url = {http://aclweb.org/anthology/N18-2002},
	doi = {10.18653/v1/N18-2002},
	language = {en},
	urldate = {2024-10-22},
	booktitle = {Proceedings of the 2018 {Conference} of the {North} {American} {Chapter} of           the {Association} for {Computational} {Linguistics}: {Human} {Language}           {Technologies}, {Volume} 2 ({Short} {Papers})},
	publisher = {Association for Computational Linguistics},
	author = {Rudinger, Rachel and Naradowsky, Jason and Leonard, Brian and Van Durme, Benjamin},
	year = {2018},
	pages = {8--14},
}

@article{blodgett_language_2020,
	title = {Language ({Technology}) is {Power}: {A} {Critical} {Survey} of "{Bias}" in {NLP}},
	shorttitle = {Language ({Technology}) is {Power}},
	url = {http://arxiv.org/abs/2005.14050},
	language = {en},
	urldate = {2024-11-13},
	archivePrefix={arXiv},
	author = {Blodgett, Su Lin and Barocas, Solon and III, Hal Daumé and Wallach, Hanna},
	month = may,
	year = {2020},
	journal={arXiv preprint arXiv:2005.14050 },
}

@article{tamkin_evaluating_2023,
	title = {Evaluating and {Mitigating} {Discrimination} in {Language} {Model} {Decisions}},
	url = {http://arxiv.org/abs/2312.03689},
	language = {en},
	urldate = {2024-11-15},
	archivePrefix={arXiv},
	author = {Tamkin, Alex and Askell, Amanda and Lovitt, Liane and Durmus, Esin and Joseph, Nicholas and Kravec, Shauna and Nguyen, Karina and Kaplan, Jared and Ganguli, Deep},
	month = dec,
	year = {2023},
	journal = {arXiv preprint arXiv:2312.03689 },
}

@inproceedings{pikuliak_women_2024,
    title = "Women Are Beautiful, Men Are Leaders: Gender Stereotypes in Machine Translation and Language Modeling",
    author = "Pikuliak, Mat{\'u}{\v{s}}  and
      Oresko, Stefan  and
      Hrckova, Andrea  and
      Simko, Marian",
    editor = "Al-Onaizan, Yaser  and
      Bansal, Mohit  and
      Chen, Yun-Nung",
    booktitle = "Findings of the Association for Computational Linguistics: EMNLP 2024",
    month = nov,
    year = "2024",
    address = "Miami, Florida, USA",
    publisher = "Association for Computational Linguistics",
    url = "https://aclanthology.org/2024.findings-emnlp.173/",
    doi = "10.18653/v1/2024.findings-emnlp.173",
    pages = "3060--3083"
}

@article{savoldi-etal-2021-gender,
    title = "Gender Bias in Machine Translation",
    author = "Savoldi, Beatrice  and
      Gaido, Marco  and
      Bentivogli, Luisa  and
      Negri, Matteo  and
      Turchi, Marco",
    editor = "Roark, Brian  and
      Nenkova, Ani",
    journal = "Transactions of the Association for Computational Linguistics",
    volume = "9",
    year = "2021",
    address = "Cambridge, MA",
    publisher = "MIT Press",
    url = "https://aclanthology.org/2021.tacl-1.51/",
    doi = "10.1162/tacl\_a\_00401",
    pages = "845--874"
}

@inproceedings{talat_you_2022,
	address = {virtual+Dublin},
	title = {You reap what you sow: {On} the {Challenges} of {Bias} {Evaluation} {Under} {Multilingual} {Settings}},
	shorttitle = {You reap what you sow},
	url = {https://aclanthology.org/2022.bigscience-1.3},
	doi = {10.18653/v1/2022.bigscience-1.3},
	language = {en},
	urldate = {2024-11-22},
	booktitle = {Proceedings of {BigScience} {Episode} \#5 -- {Workshop} on {Challenges} \& {Perspectives} in {Creating} {Large} {Language} {Models}},
	publisher = {Association for Computational Linguistics},
	author = {Talat, Zeerak and Névéol, Aurélie and Biderman, Stella and Clinciu, Miruna and Dey, Manan and Longpre, Shayne and Luccioni, Sasha and Masoud, Maraim and Mitchell, Margaret and Radev, Dragomir and Sharma, Shanya and Subramonian, Arjun and Tae, Jaesung and Tan, Samson and Tunuguntla, Deepak and Van Der Wal, Oskar},
	year = {2022},
	pages = {26--41},
}

@inproceedings{wan_kelly_2023,
	address = {Singapore},
	title = {“{Kelly} is a {Warm} {Person}, {Joseph} is a {Role} {Model}”: {Gender} {Biases} in {LLM}-{Generated} {Reference} {Letters}},
	shorttitle = {“{Kelly} is a {Warm} {Person}, {Joseph} is a {Role} {Model}”},
	url = {https://aclanthology.org/2023.findings-emnlp.243},
	doi = {10.18653/v1/2023.findings-emnlp.243},
	language = {en},
	urldate = {2024-11-22},
	booktitle = {Findings of the {Association} for {Computational} {Linguistics}: {EMNLP} 2023},
	publisher = {Association for Computational Linguistics},
	author = {Wan, Yixin and Pu, George and Sun, Jiao and Garimella, Aparna and Chang, Kai-Wei and Peng, Nanyun},
	year = {2023},
	pages = {3730--3748},
}

@inproceedings{stanovsky_evaluating_2019,
	address = {Florence, Italy},
	title = {Evaluating {Gender} {Bias} in {Machine} {Translation}},
	url = {https://www.aclweb.org/anthology/P19-1164},
	doi = {10.18653/v1/P19-1164},
	language = {en},
	urldate = {2024-12-02},
	booktitle = {Proceedings of the 57th {Annual} {Meeting} of the {Association} for {Computational} {Linguistics}},
	publisher = {Association for Computational Linguistics},
	author = {Stanovsky, Gabriel and Smith, Noah A. and Zettlemoyer, Luke},
	year = {2019},
	pages = {1679--1684},
}

@inproceedings{mitchell-etal-2025-shades,
    title = "{SHADES}: Towards a Multilingual Assessment of Stereotypes in Large Language Models",
    author = "Mitchell, Margaret  and
      Attanasio, Giuseppe  and
      Baldini, Ioana  and
      Clinciu, Miruna  and
      Clive, Jordan  and
      Delobelle, Pieter  and
      Dey, Manan  and
      Hamilton, Sil  and
      Dill, Timm  and
      Doughman, Jad  and
      Dutt, Ritam  and
      Ghosh, Avijit  and
      Forde, Jessica Zosa  and
      Holtermann, Carolin  and
      Kaffee, Lucie-Aim{\'e}e  and
      Laud, Tanmay  and
      Lauscher, Anne  and
      Lopez-Davila, Roberto L  and
      Masoud, Maraim  and
      Nangia, Nikita  and
      Ovalle, Anaelia  and
      Pistilli, Giada  and
      Radev, Dragomir  and
      Savoldi, Beatrice  and
      Raheja, Vipul  and
      Qin, Jeremy  and
      Ploeger, Esther  and
      Subramonian, Arjun  and
      Dhole, Kaustubh  and
      Sun, Kaiser  and
      Djanibekov, Amirbek  and
      Mansurov, Jonibek  and
      Yin, Kayo  and
      Cueva, Emilio Villa  and
      Mukherjee, Sagnik  and
      Huang, Jerry  and
      Shen, Xudong  and
      Gala, Jay  and
      Al-Ali, Hamdan  and
      Tair Djanibekov  and
      Mukhituly, Nurdaulet  and
      Nie, Shangrui  and
      Sharma, Shanya  and
      Stanczak, Karolina  and
      Szczechla, Eliza  and
      Timponi Torrent, Tiago  and
      Tunuguntla, Deepak  and
      Viridiano, Marcelo  and
      Van Der Wal, Oskar  and
      Yakefu, Adina  and
      N{\'e}v{\'e}ol, Aur{\'e}lie  and
      Zhang, Mike  and
      Zink, Sydney  and
      Talat, Zeerak",
    editor = "Chiruzzo, Luis  and
      Ritter, Alan  and
      Wang, Lu",
    booktitle = "Proceedings of the 2025 Conference of the Nations of the Americas Chapter of the Association for Computational Linguistics: Human Language Technologies (Volume 1: Long Papers)",
    month = apr,
    year = "2025",
    address = "Albuquerque, New Mexico",
    publisher = "Association for Computational Linguistics",
    url = "https://aclanthology.org/2025.naacl-long.600/",
    pages = "11995--12041",
    ISBN = "979-8-89176-189-6"
}

@inproceedings{dev_harms_2021,
	address = {Online and Punta Cana, Dominican Republic},
	title = {Harms of {Gender} {Exclusivity} and {Challenges} in {Non}-{Binary} {Representation} in {Language} {Technologies}},
	url = {https://aclanthology.org/2021.emnlp-main.150},
	doi = {10.18653/v1/2021.emnlp-main.150},
	language = {en},
	urldate = {2024-12-02},
	booktitle = {Proceedings of the 2021 {Conference} on {Empirical} {Methods} in {Natural} {Language} {Processing}},
	publisher = {Association for Computational Linguistics},
	author = {Dev, Sunipa and Monajatipoor, Masoud and Ovalle, Anaelia and Subramonian, Arjun and Phillips, Jeff and Chang, Kai-Wei},
	year = {2021},
	pages = {1968--1994},
}

@inproceedings{devbuilding,
author = {Dev, Sunipa and Goyal, Jaya and Tewari, Dinesh and Dave, Shachi and Prabhakaran, Vinodkumar},
title = {Building socio-culturally inclusive stereotype resources with community engagement},
year = {2023},
publisher = {Curran Associates Inc.},
address = {Red Hook, NY, USA},
booktitle = {Proceedings of the 37th International Conference on Neural Information Processing Systems},
articleno = {194},
numpages = {17},
location = {New Orleans, LA, USA},
series = {NIPS '23},
url = {https://proceedings.neurips.cc/paper\_files/paper/2023/file/0dc91de822b71c66a7f54fa121d8cbb9-Paper-Datasets\_and\_Benchmarks.pdf}
}

@article{gallegos_bias_2024,
	title = {Bias and {Fairness} in {Large} {Language} {Models}: {A} {Survey}},
	volume = {50},
	issn = {0891-2017, 1530-9312},
	shorttitle = {Bias and {Fairness} in {Large} {Language} {Models}},
	url = {https://direct.mit.edu/coli/article/50/3/1097/121961/Bias-and-Fairness-in-Large-Language-Models-A},
	doi = {10.1162/coli\_a\_00524},
	language = {en},
	number = {3},
	urldate = {2024-12-04},
	journal = {Computational Linguistics},
	author = {Gallegos, Isabel O. and Rossi, Ryan A. and Barrow, Joe and Tanjim, Md Mehrab and Kim, Sungchul and Dernoncourt, Franck and Yu, Tong and Zhang, Ruiyi and Ahmed, Nesreen K.},
	month = sep,
	year = {2024},
	pages = {1097--1179},
}

@article{bartl_gender_2025,
	title = {Gender {Bias} in {Natural} {Language} {Processing} and {Computer} {Vision}: {A} {Comparative} {Survey}},
	volume = {57},
	issn = {0360-0300, 1557-7341},
	shorttitle = {Gender {Bias} in {Natural} {Language} {Processing} and {Computer} {Vision}},
	url = {https://dl.acm.org/doi/10.1145/3700438},
	doi = {10.1145/3700438},
	language = {en},
	number = {6},
	urldate = {2025-02-25},
	journal = {ACM Computing Surveys},
	author = {Bartl, Marion and Mandal, Abhishek and Leavy, Susan and Little, Suzanne},
	month = jun,
	year = {2025},
	pages = {1--36},
}

@misc{ranjan_comprehensive_2024,
	title = {A {Comprehensive} {Survey} of {Bias} in {LLMs}: {Current} {Landscape} and {Future} {Directions}},
	shorttitle = {A {Comprehensive} {Survey} of {Bias} in {LLMs}},
	url = {http://arxiv.org/abs/2409.16430},
	doi = {10.48550/arXiv.2409.16430},
	urldate = {2025-04-15},
	archivePrefix={arXiv},
	author = {Ranjan, Rajesh and Gupta, Shailja and Singh, Surya Narayan},
	month = sep,
	year = {2024},
	note = {arXiv:2409.16430 },
}

@article{glaese_improving_2022,
	title = {Improving alignment of dialogue agents via targeted human judgements},
	url = {http://arxiv.org/abs/2209.14375},
	doi = {10.48550/arXiv.2209.14375},
	language = {en},
	urldate = {2025-04-15},
	archivePrefix={arXiv},
	author = {Glaese, Amelia and McAleese, Nat and Trębacz, Maja and Aslanides, John and Firoiu, Vlad and Ewalds, Timo and Rauh, Maribeth and Weidinger, Laura and Chadwick, Martin and Thacker, Phoebe and Campbell-Gillingham, Lucy and Uesato, Jonathan and Huang, Po-Sen and Comanescu, Ramona and Yang, Fan and See, Abigail and Dathathri, Sumanth and Greig, Rory and Chen, Charlie and Fritz, Doug and Elias, Jaume Sanchez and Green, Richard and Mokrá, Soňa and Fernando, Nicholas and Wu, Boxi and Foley, Rachel and Young, Susannah and Gabriel, Iason and Isaac, William and Mellor, John and Hassabis, Demis and Kavukcuoglu, Koray and Hendricks, Lisa Anne and Irving, Geoffrey},
	month = sep,
	year = {2022},
	journal={arXiv preprint arXiv:2209.14375 },
}

@inproceedings{munro_detecting_2020,
	address = {Online},
	title = {Detecting {Independent} {Pronoun} {Bias} with {Partially}-{Synthetic} {Data} {Generation}},
	url = {https://www.aclweb.org/anthology/2020.emnlp-main.157},
	doi = {10.18653/v1/2020.emnlp-main.157},
	language = {en},
	urldate = {2025-04-15},
	booktitle = {Proceedings of the 2020 {Conference} on {Empirical} {Methods} in {Natural} {Language} {Processing} ({EMNLP})},
	publisher = {Association for Computational Linguistics},
	author = {Munro, Robert and Morrison, Alex (Carmen)},
	year = {2020},
	pages = {2011--2017},
}

@inproceedings{shelby_sociotechnical_2023,
	address = {Montréal QC Canada},
	title = {Sociotechnical {Harms} of {Algorithmic} {Systems}: {Scoping} a {Taxonomy} for {Harm} {Reduction}},
	isbn = {9798400702310},
	shorttitle = {Sociotechnical {Harms} of {Algorithmic} {Systems}},
	url = {https://dl.acm.org/doi/10.1145/3600211.3604673},
	doi = {10.1145/3600211.3604673},
	language = {en},
	urldate = {2025-04-15},
	booktitle = {Proceedings of the 2023 {AAAI}/{ACM} {Conference} on {AI}, {Ethics}, and {Society}},
	publisher = {ACM},
	author = {Shelby, Renee and Rismani, Shalaleh and Henne, Kathryn and Moon, AJung and Rostamzadeh, Negar and Nicholas, Paul and Yilla-Akbari, N'Mah and Gallegos, Jess and Smart, Andrew and Garcia, Emilio and Virk, Gurleen},
	month = aug,
	year = {2023},
	pages = {723--741},
}

@article{stanczak_survey_2021,
	title = {A {Survey} on {Gender} {Bias} in {Natural} {Language} {Processing}},
	url = {http://arxiv.org/abs/2112.14168},
	doi = {10.48550/arXiv.2112.14168},
	language = {en},
	urldate = {2025-04-15},
	archivePrefix={arXiv},
	author = {Stanczak, Karolina and Augenstein, Isabelle},
	month = dec,
	year = {2021},
	journal={arXiv preprint arXiv:2112.14168},
}

@misc{rei_comet_2020,
	title = {{COMET}: {A} {Neural} {Framework} for {MT} {Evaluation}},
	shorttitle = {{COMET}},
	url = {http://arxiv.org/abs/2009.09025},
	doi = {10.48550/arXiv.2009.09025},
	language = {en},
	urldate = {2025-04-16},
	archivePrefix={arXiv},
	author = {Rei, Ricardo and Stewart, Craig and Farinha, Ana C. and Lavie, Alon},
	month = oct,
	year = {2020},
	note = {arXiv:2009.09025 },
}

@article{kocmi_error_2024,
	title = {Error {Span} {Annotation}: {A} {Balanced} {Approach} for {Human} {Evaluation} of {Machine} {Translation}},
	shorttitle = {Error {Span} {Annotation}},
	url = {http://arxiv.org/abs/2406.11580},
	doi = {10.48550/arXiv.2406.11580},
	language = {en},
	urldate = {2025-04-16},
	archivePrefix={arXiv},
	author = {Kocmi, Tom and Zouhar, Vilém and Avramidis, Eleftherios and Grundkiewicz, Roman and Karpinska, Marzena and Popović, Maja and Sachan, Mrinmaya and Shmatova, Mariya},
	month = oct,
	year = {2024},
	journal={arXiv preprint arXiv:2406.11580 },
}

@article{geminiteam2024gemini15unlockingmultimodal,
      title={Gemini 1.5: Unlocking multimodal understanding across millions of tokens of context}, 
      author={{Gemini Team} and Petko Georgiev and Ving Ian Lei and Ryan Burnell and Libin Bai and Anmol Gulati and Garrett Tanzer and Damien Vincent and Zhufeng Pan and Shibo Wang and Soroosh Mariooryad and Yifan Ding and Xinyang Geng and Fred Alcober and Roy Frostig and Mark Omernick and Lexi Walker and Cosmin Paduraru and Christina Sorokin and Andrea Tacchetti and Colin Gaffney and Samira Daruki and Olcan Sercinoglu and Zach Gleicher and Juliette Love and Paul Voigtlaender and Rohan Jain and Gabriela Surita and Kareem Mohamed and Rory Blevins and Junwhan Ahn and Tao Zhu and Kornraphop Kawintiranon and Orhan Firat and Yiming Gu and Yujing Zhang and Matthew Rahtz and Manaal Faruqui and Natalie Clay and Justin Gilmer and JD Co-Reyes and Ivo Penchev and Rui Zhu and Nobuyuki Morioka and Kevin Hui and Krishna Haridasan and Victor Campos and Mahdis Mahdieh and Mandy Guo and Samer Hassan and Kevin Kilgour and Arpi Vezer and Heng-Tze Cheng and Raoul de Liedekerke and Siddharth Goyal and Paul Barham and DJ Strouse and Seb Noury and Jonas Adler and Mukund Sundararajan and Sharad Vikram and Dmitry Lepikhin and Michela Paganini and Xavier Garcia and Fan Yang and Dasha Valter and Maja Trebacz and Kiran Vodrahalli and Chulayuth Asawaroengchai and Roman Ring and Norbert Kalb and Livio Baldini Soares and Siddhartha Brahma and David Steiner and Tianhe Yu and Fabian Mentzer and Antoine He and Lucas Gonzalez and Bibo Xu and Raphael Lopez Kaufman and Laurent El Shafey and Junhyuk Oh and Tom Hennigan and George van den Driessche and Seth Odoom and Mario Lucic and Becca Roelofs and Sid Lall and Amit Marathe and Betty Chan and Santiago Ontanon and Luheng He and Denis Teplyashin and Jonathan Lai and Phil Crone and Bogdan Damoc and Lewis Ho and Sebastian Riedel and Karel Lenc and Chih-Kuan Yeh and Aakanksha Chowdhery and Yang Xu and Mehran Kazemi and Ehsan Amid and Anastasia Petrushkina and Kevin Swersky and Ali Khodaei and Gowoon Chen and Chris Larkin and Mario Pinto and Geng Yan and Adria Puigdomenech Badia and Piyush Patil and Steven Hansen and Dave Orr and Sebastien M. R. Arnold and Jordan Grimstad and Andrew Dai and Sholto Douglas and Rishika Sinha and Vikas Yadav and Xi Chen and Elena Gribovskaya and Jacob Austin and Jeffrey Zhao and Kaushal Patel and Paul Komarek and Sophia Austin and Sebastian Borgeaud and Linda Friso and Abhimanyu Goyal and Ben Caine and Kris Cao and Da-Woon Chung and Matthew Lamm and Gabe Barth-Maron and Thais Kagohara and Kate Olszewska and Mia Chen and Kaushik Shivakumar and Rishabh Agarwal and Harshal Godhia and Ravi Rajwar and Javier Snaider and Xerxes Dotiwalla and Yuan Liu and Aditya Barua and Victor Ungureanu and Yuan Zhang and Bat-Orgil Batsaikhan and Mateo Wirth and James Qin and Ivo Danihelka and Tulsee Doshi and Martin Chadwick and Jilin Chen and Sanil Jain and Quoc Le and Arjun Kar and Madhu Gurumurthy and Cheng Li and Ruoxin Sang and Fangyu Liu and Lampros Lamprou and Rich Munoz and Nathan Lintz and Harsh Mehta and Heidi Howard and Malcolm Reynolds and Lora Aroyo and Quan Wang and Lorenzo Blanco and Albin Cassirer and Jordan Griffith and Dipanjan Das and Stephan Lee and Jakub Sygnowski and Zach Fisher and James Besley and Richard Powell and Zafarali Ahmed and Dominik Paulus and David Reitter and Zalan Borsos and Rishabh Joshi and Aedan Pope and Steven Hand and Vittorio Selo and Vihan Jain and Nikhil Sethi and Megha Goel and Takaki Makino and Rhys May and Zhen Yang and Johan Schalkwyk and Christina Butterfield and Anja Hauth and Alex Goldin and Will Hawkins and Evan Senter and Sergey Brin and Oliver Woodman and Marvin Ritter and Eric Noland and Minh Giang and Vijay Bolina and Lisa Lee and Tim Blyth and Ian Mackinnon and Machel Reid and Obaid Sarvana and David Silver and Alexander Chen and Lily Wang and Loren Maggiore and Oscar Chang and Nithya Attaluri and Gregory Thornton and Chung-Cheng Chiu and Oskar Bunyan and Nir Levine and Timothy Chung and Evgenii Eltyshev and Xiance Si and Timothy Lillicrap and Demetra Brady and Vaibhav Aggarwal and Boxi Wu and Yuanzhong Xu and Ross McIlroy and Kartikeya Badola and Paramjit Sandhu and Erica Moreira and Wojciech Stokowiec and Ross Hemsley and Dong Li and Alex Tudor and Pranav Shyam and Elahe Rahimtoroghi and Salem Haykal and Pablo Sprechmann and Xiang Zhou and Diana Mincu and Yujia Li and Ravi Addanki and Kalpesh Krishna and Xiao Wu and Alexandre Frechette and Matan Eyal and Allan Dafoe and Dave Lacey and Jay Whang and Thi Avrahami and Ye Zhang and Emanuel Taropa and Hanzhao Lin and Daniel Toyama and Eliza Rutherford and Motoki Sano and HyunJeong Choe and Alex Tomala and Chalence Safranek-Shrader and Nora Kassner and Mantas Pajarskas and Matt Harvey and Sean Sechrist and Meire Fortunato and Christina Lyu and Gamaleldin Elsayed and Chenkai Kuang and James Lottes and Eric Chu and Chao Jia and Chih-Wei Chen and Peter Humphreys and Kate Baumli and Connie Tao and Rajkumar Samuel and Cicero Nogueira dos Santos and Anders Andreassen and Nemanja Rakićević and Dominik Grewe and Aviral Kumar and Stephanie Winkler and Jonathan Caton and Andrew Brock and Sid Dalmia and Hannah Sheahan and Iain Barr and Yingjie Miao and Paul Natsev and Jacob Devlin and Feryal Behbahani and Flavien Prost and Yanhua Sun and Artiom Myaskovsky and Thanumalayan Sankaranarayana Pillai and Dan Hurt and Angeliki Lazaridou and Xi Xiong and Ce Zheng and Fabio Pardo and Xiaowei Li and Dan Horgan and Joe Stanton and Moran Ambar and Fei Xia and Alejandro Lince and Mingqiu Wang and Basil Mustafa and Albert Webson and Hyo Lee and Rohan Anil and Martin Wicke and Timothy Dozat and Abhishek Sinha and Enrique Piqueras and Elahe Dabir and Shyam Upadhyay and Anudhyan Boral and Lisa Anne Hendricks and Corey Fry and Josip Djolonga and Yi Su and Jake Walker and Jane Labanowski and Ronny Huang and Vedant Misra and Jeremy Chen and RJ Skerry-Ryan and Avi Singh and Shruti Rijhwani and Dian Yu and Alex Castro-Ros and Beer Changpinyo and Romina Datta and Sumit Bagri and Arnar Mar Hrafnkelsson and Marcello Maggioni and Daniel Zheng and Yury Sulsky and Shaobo Hou and Tom Le Paine and Antoine Yang and Jason Riesa and Dominika Rogozinska and Dror Marcus and Dalia El Badawy and Qiao Zhang and Luyu Wang and Helen Miller and Jeremy Greer and Lars Lowe Sjos and Azade Nova and Heiga Zen and Rahma Chaabouni and Mihaela Rosca and Jiepu Jiang and Charlie Chen and Ruibo Liu and Tara Sainath and Maxim Krikun and Alex Polozov and Jean-Baptiste Lespiau and Josh Newlan and Zeyncep Cankara and Soo Kwak and Yunhan Xu and Phil Chen and Andy Coenen and Clemens Meyer and Katerina Tsihlas and Ada Ma and Juraj Gottweis and Jinwei Xing and Chenjie Gu and Jin Miao and Christian Frank and Zeynep Cankara and Sanjay Ganapathy and Ishita Dasgupta and Steph Hughes-Fitt and Heng Chen and David Reid and Keran Rong and Hongmin Fan and Joost van Amersfoort and Vincent Zhuang and Aaron Cohen and Shixiang Shane Gu and Anhad Mohananey and Anastasija Ilic and Taylor Tobin and John Wieting and Anna Bortsova and Phoebe Thacker and Emma Wang and Emily Caveness and Justin Chiu and Eren Sezener and Alex Kaskasoli and Steven Baker and Katie Millican and Mohamed Elhawaty and Kostas Aisopos and Carl Lebsack and Nathan Byrd and Hanjun Dai and Wenhao Jia and Matthew Wiethoff and Elnaz Davoodi and Albert Weston and Lakshman Yagati and Arun Ahuja and Isabel Gao and Golan Pundak and Susan Zhang and Michael Azzam and Khe Chai Sim and Sergi Caelles and James Keeling and Abhanshu Sharma and Andy Swing and YaGuang Li and Chenxi Liu and Carrie Grimes Bostock and Yamini Bansal and Zachary Nado and Ankesh Anand and Josh Lipschultz and Abhijit Karmarkar and Lev Proleev and Abe Ittycheriah and Soheil Hassas Yeganeh and George Polovets and Aleksandra Faust and Jiao Sun and Alban Rrustemi and Pen Li and Rakesh Shivanna and Jeremiah Liu and Chris Welty and Federico Lebron and Anirudh Baddepudi and Sebastian Krause and Emilio Parisotto and Radu Soricut and Zheng Xu and Dawn Bloxwich and Melvin Johnson and Behnam Neyshabur and Justin Mao-Jones and Renshen Wang and Vinay Ramasesh and Zaheer Abbas and Arthur Guez and Constant Segal and Duc Dung Nguyen and James Svensson and Le Hou and Sarah York and Kieran Milan and Sophie Bridgers and Wiktor Gworek and Marco Tagliasacchi and James Lee-Thorp and Michael Chang and Alexey Guseynov and Ale Jakse Hartman and Michael Kwong and Ruizhe Zhao and Sheleem Kashem and Elizabeth Cole and Antoine Miech and Richard Tanburn and Mary Phuong and Filip Pavetic and Sebastien Cevey and Ramona Comanescu and Richard Ives and Sherry Yang and Cosmo Du and Bo Li and Zizhao Zhang and Mariko Iinuma and Clara Huiyi Hu and Aurko Roy and Shaan Bijwadia and Zhenkai Zhu and Danilo Martins and Rachel Saputro and Anita Gergely and Steven Zheng and Dawei Jia and Ioannis Antonoglou and Adam Sadovsky and Shane Gu and Yingying Bi and Alek Andreev and Sina Samangooei and Mina Khan and Tomas Kocisky and Angelos Filos and Chintu Kumar and Colton Bishop and Adams Yu and Sarah Hodkinson and Sid Mittal and Premal Shah and Alexandre Moufarek and Yong Cheng and Adam Bloniarz and Jaehoon Lee and Pedram Pejman and Paul Michel and Stephen Spencer and Vladimir Feinberg and Xuehan Xiong and Nikolay Savinov and Charlotte Smith and Siamak Shakeri and Dustin Tran and Mary Chesus and Bernd Bohnet and George Tucker and Tamara von Glehn and Carrie Muir and Yiran Mao and Hideto Kazawa and Ambrose Slone and Kedar Soparkar and Disha Shrivastava and James Cobon-Kerr and Michael Sharman and Jay Pavagadhi and Carlos Araya and Karolis Misiunas and Nimesh Ghelani and Michael Laskin and David Barker and Qiujia Li and Anton Briukhov and Neil Houlsby and Mia Glaese and Balaji Lakshminarayanan and Nathan Schucher and Yunhao Tang and Eli Collins and Hyeontaek Lim and Fangxiaoyu Feng and Adria Recasens and Guangda Lai and Alberto Magni and Nicola De Cao and Aditya Siddhant and Zoe Ashwood and Jordi Orbay and Mostafa Dehghani and Jenny Brennan and Yifan He and Kelvin Xu and Yang Gao and Carl Saroufim and James Molloy and Xinyi Wu and Seb Arnold and Solomon Chang and Julian Schrittwieser and Elena Buchatskaya and Soroush Radpour and Martin Polacek and Skye Giordano and Ankur Bapna and Simon Tokumine and Vincent Hellendoorn and Thibault Sottiaux and Sarah Cogan and Aliaksei Severyn and Mohammad Saleh and Shantanu Thakoor and Laurent Shefey and Siyuan Qiao and Meenu Gaba and Shuo-yiin Chang and Craig Swanson and Biao Zhang and Benjamin Lee and Paul Kishan Rubenstein and Gan Song and Tom Kwiatkowski and Anna Koop and Ajay Kannan and David Kao and Parker Schuh and Axel Stjerngren and Golnaz Ghiasi and Gena Gibson and Luke Vilnis and Ye Yuan and Felipe Tiengo Ferreira and Aishwarya Kamath and Ted Klimenko and Ken Franko and Kefan Xiao and Indro Bhattacharya and Miteyan Patel and Rui Wang and Alex Morris and Robin Strudel and Vivek Sharma and Peter Choy and Sayed Hadi Hashemi and Jessica Landon and Mara Finkelstein and Priya Jhakra and Justin Frye and Megan Barnes and Matthew Mauger and Dennis Daun and Khuslen Baatarsukh and Matthew Tung and Wael Farhan and Henryk Michalewski and Fabio Viola and Felix de Chaumont Quitry and Charline Le Lan and Tom Hudson and Qingze Wang and Felix Fischer and Ivy Zheng and Elspeth White and Anca Dragan and Jean-baptiste Alayrac and Eric Ni and Alexander Pritzel and Adam Iwanicki and Michael Isard and Anna Bulanova and Lukas Zilka and Ethan Dyer and Devendra Sachan and Srivatsan Srinivasan and Hannah Muckenhirn and Honglong Cai and Amol Mandhane and Mukarram Tariq and Jack W. Rae and Gary Wang and Kareem Ayoub and Nicholas FitzGerald and Yao Zhao and Woohyun Han and Chris Alberti and Dan Garrette and Kashyap Krishnakumar and Mai Gimenez and Anselm Levskaya and Daniel Sohn and Josip Matak and Inaki Iturrate and Michael B. Chang and Jackie Xiang and Yuan Cao and Nishant Ranka and Geoff Brown and Adrian Hutter and Vahab Mirrokni and Nanxin Chen and Kaisheng Yao and Zoltan Egyed and Francois Galilee and Tyler Liechty and Praveen Kallakuri and Evan Palmer and Sanjay Ghemawat and Jasmine Liu and David Tao and Chloe Thornton and Tim Green and Mimi Jasarevic and Sharon Lin and Victor Cotruta and Yi-Xuan Tan and Noah Fiedel and Hongkun Yu and Ed Chi and Alexander Neitz and Jens Heitkaemper and Anu Sinha and Denny Zhou and Yi Sun and Charbel Kaed and Brice Hulse and Swaroop Mishra and Maria Georgaki and Sneha Kudugunta and Clement Farabet and Izhak Shafran and Daniel Vlasic and Anton Tsitsulin and Rajagopal Ananthanarayanan and Alen Carin and Guolong Su and Pei Sun and Shashank V and Gabriel Carvajal and Josef Broder and Iulia Comsa and Alena Repina and William Wong and Warren Weilun Chen and Peter Hawkins and Egor Filonov and Lucia Loher and Christoph Hirnschall and Weiyi Wang and Jingchen Ye and Andrea Burns and Hardie Cate and Diana Gage Wright and Federico Piccinini and Lei Zhang and Chu-Cheng Lin and Ionel Gog and Yana Kulizhskaya and Ashwin Sreevatsa and Shuang Song and Luis C. Cobo and Anand Iyer and Chetan Tekur and Guillermo Garrido and Zhuyun Xiao and Rupert Kemp and Huaixiu Steven Zheng and Hui Li and Ananth Agarwal and Christel Ngani and Kati Goshvadi and Rebeca Santamaria-Fernandez and Wojciech Fica and Xinyun Chen and Chris Gorgolewski and Sean Sun and Roopal Garg and Xinyu Ye and S. M. Ali Eslami and Nan Hua and Jon Simon and Pratik Joshi and Yelin Kim and Ian Tenney and Sahitya Potluri and Lam Nguyen Thiet and Quan Yuan and Florian Luisier and Alexandra Chronopoulou and Salvatore Scellato and Praveen Srinivasan and Minmin Chen and Vinod Koverkathu and Valentin Dalibard and Yaming Xu and Brennan Saeta and Keith Anderson and Thibault Sellam and Nick Fernando and Fantine Huot and Junehyuk Jung and Mani Varadarajan and Michael Quinn and Amit Raul and Maigo Le and Ruslan Habalov and Jon Clark and Komal Jalan and Kalesha Bullard and Achintya Singhal and Thang Luong and Boyu Wang and Sujeevan Rajayogam and Julian Eisenschlos and Johnson Jia and Daniel Finchelstein and Alex Yakubovich and Daniel Balle and Michael Fink and Sameer Agarwal and Jing Li and Dj Dvijotham and Shalini Pal and Kai Kang and Jaclyn Konzelmann and Jennifer Beattie and Olivier Dousse and Diane Wu and Remi Crocker and Chen Elkind and Siddhartha Reddy Jonnalagadda and Jong Lee and Dan Holtmann-Rice and Krystal Kallarackal and Rosanne Liu and Denis Vnukov and Neera Vats and Luca Invernizzi and Mohsen Jafari and Huanjie Zhou and Lilly Taylor and Jennifer Prendki and Marcus Wu and Tom Eccles and Tianqi Liu and Kavya Kopparapu and Francoise Beaufays and Christof Angermueller and Andreea Marzoca and Shourya Sarcar and Hilal Dib and Jeff Stanway and Frank Perbet and Nejc Trdin and Rachel Sterneck and Andrey Khorlin and Dinghua Li and Xihui Wu and Sonam Goenka and David Madras and Sasha Goldshtein and Willi Gierke and Tong Zhou and Yaxin Liu and Yannie Liang and Anais White and Yunjie Li and Shreya Singh and Sanaz Bahargam and Mark Epstein and Sujoy Basu and Li Lao and Adnan Ozturel and Carl Crous and Alex Zhai and Han Lu and Zora Tung and Neeraj Gaur and Alanna Walton and Lucas Dixon and Ming Zhang and Amir Globerson and Grant Uy and Andrew Bolt and Olivia Wiles and Milad Nasr and Ilia Shumailov and Marco Selvi and Francesco Piccinno and Ricardo Aguilar and Sara McCarthy and Misha Khalman and Mrinal Shukla and Vlado Galic and John Carpenter and Kevin Villela and Haibin Zhang and Harry Richardson and James Martens and Matko Bosnjak and Shreyas Rammohan Belle and Jeff Seibert and Mahmoud Alnahlawi and Brian McWilliams and Sankalp Singh and Annie Louis and Wen Ding and Dan Popovici and Lenin Simicich and Laura Knight and Pulkit Mehta and Nishesh Gupta and Chongyang Shi and Saaber Fatehi and Jovana Mitrovic and Alex Grills and Joseph Pagadora and Tsendsuren Munkhdalai and Dessie Petrova and Danielle Eisenbud and Zhishuai Zhang and Damion Yates and Bhavishya Mittal and Nilesh Tripuraneni and Yannis Assael and Thomas Brovelli and Prateek Jain and Mihajlo Velimirovic and Canfer Akbulut and Jiaqi Mu and Wolfgang Macherey and Ravin Kumar and Jun Xu and Haroon Qureshi and Gheorghe Comanici and Jeremy Wiesner and Zhitao Gong and Anton Ruddock and Matthias Bauer and Nick Felt and Anirudh GP and Anurag Arnab and Dustin Zelle and Jonas Rothfuss and Bill Rosgen and Ashish Shenoy and Bryan Seybold and Xinjian Li and Jayaram Mudigonda and Goker Erdogan and Jiawei Xia and Jiri Simsa and Andrea Michi and Yi Yao and Christopher Yew and Steven Kan and Isaac Caswell and Carey Radebaugh and Andre Elisseeff and Pedro Valenzuela and Kay McKinney and Kim Paterson and Albert Cui and Eri Latorre-Chimoto and Solomon Kim and William Zeng and Ken Durden and Priya Ponnapalli and Tiberiu Sosea and Christopher A. Choquette-Choo and James Manyika and Brona Robenek and Harsha Vashisht and Sebastien Pereira and Hoi Lam and Marko Velic and Denese Owusu-Afriyie and Katherine Lee and Tolga Bolukbasi and Alicia Parrish and Shawn Lu and Jane Park and Balaji Venkatraman and Alice Talbert and Lambert Rosique and Yuchung Cheng and Andrei Sozanschi and Adam Paszke and Praveen Kumar and Jessica Austin and Lu Li and Khalid Salama and Bartek Perz and Wooyeol Kim and Nandita Dukkipati and Anthony Baryshnikov and Christos Kaplanis and XiangHai Sheng and Yuri Chervonyi and Caglar Unlu and Diego de Las Casas and Harry Askham and Kathryn Tunyasuvunakool and Felix Gimeno and Siim Poder and Chester Kwak and Matt Miecnikowski and Vahab Mirrokni and Alek Dimitriev and Aaron Parisi and Dangyi Liu and Tomy Tsai and Toby Shevlane and Christina Kouridi and Drew Garmon and Adrian Goedeckemeyer and Adam R. Brown and Anitha Vijayakumar and Ali Elqursh and Sadegh Jazayeri and Jin Huang and Sara Mc Carthy and Jay Hoover and Lucy Kim and Sandeep Kumar and Wei Chen and Courtney Biles and Garrett Bingham and Evan Rosen and Lisa Wang and Qijun Tan and David Engel and Francesco Pongetti and Dario de Cesare and Dongseong Hwang and Lily Yu and Jennifer Pullman and Srini Narayanan and Kyle Levin and Siddharth Gopal and Megan Li and Asaf Aharoni and Trieu Trinh and Jessica Lo and Norman Casagrande and Roopali Vij and Loic Matthey and Bramandia Ramadhana and Austin Matthews and CJ Carey and Matthew Johnson and Kremena Goranova and Rohin Shah and Shereen Ashraf and Kingshuk Dasgupta and Rasmus Larsen and Yicheng Wang and Manish Reddy Vuyyuru and Chong Jiang and Joana Ijazi and Kazuki Osawa and Celine Smith and Ramya Sree Boppana and Taylan Bilal and Yuma Koizumi and Ying Xu and Yasemin Altun and Nir Shabat and Ben Bariach and Alex Korchemniy and Kiam Choo and Olaf Ronneberger and Chimezie Iwuanyanwu and Shubin Zhao and David Soergel and Cho-Jui Hsieh and Irene Cai and Shariq Iqbal and Martin Sundermeyer and Zhe Chen and Elie Bursztein and Chaitanya Malaviya and Fadi Biadsy and Prakash Shroff and Inderjit Dhillon and Tejasi Latkar and Chris Dyer and Hannah Forbes and Massimo Nicosia and Vitaly Nikolaev and Somer Greene and Marin Georgiev and Pidong Wang and Nina Martin and Hanie Sedghi and John Zhang and Praseem Banzal and Doug Fritz and Vikram Rao and Xuezhi Wang and Jiageng Zhang and Viorica Patraucean and Dayou Du and Igor Mordatch and Ivan Jurin and Lewis Liu and Ayush Dubey and Abhi Mohan and Janek Nowakowski and Vlad-Doru Ion and Nan Wei and Reiko Tojo and Maria Abi Raad and Drew A. Hudson and Vaishakh Keshava and Shubham Agrawal and Kevin Ramirez and Zhichun Wu and Hoang Nguyen and Ji Liu and Madhavi Sewak and Bryce Petrini and DongHyun Choi and Ivan Philips and Ziyue Wang and Ioana Bica and Ankush Garg and Jarek Wilkiewicz and Priyanka Agrawal and Xiaowei Li and Danhao Guo and Emily Xue and Naseer Shaik and Andrew Leach and Sadh MNM Khan and Julia Wiesinger and Sammy Jerome and Abhishek Chakladar and Alek Wenjiao Wang and Tina Ornduff and Folake Abu and Alireza Ghaffarkhah and Marcus Wainwright and Mario Cortes and Frederick Liu and Joshua Maynez and Andreas Terzis and Pouya Samangouei and Riham Mansour and Tomasz Kępa and François-Xavier Aubet and Anton Algymr and Dan Banica and Agoston Weisz and Andras Orban and Alexandre Senges and Ewa Andrejczuk and Mark Geller and Niccolo Dal Santo and Valentin Anklin and Majd Al Merey and Martin Baeuml and Trevor Strohman and Junwen Bai and Slav Petrov and Yonghui Wu and Demis Hassabis and Koray Kavukcuoglu and Jeff Dean and Oriol Vinyals},
      year={2024},
      journal={arXiv preprint arXiv:2403.05530},
      archivePrefix={arXiv},
      primaryClass={cs.CL},
      url={https://arxiv.org/abs/2403.05530}, 
}

@inproceedings{barocas_problem_2017,
  author    = {Solon Barocas and Kate Crawford and Aaron Shapiro and Hanna Wallach},
  title     = {The Problem with Bias: From Allocative to Representational Harms in Machine Learning},
  booktitle = {Special Interest Group for Computing, Information and Society (SIGCIS)},
  year      = {2017}
}

@article{nllbteam2022languageleftbehindscaling,
      title={No Language Left Behind: Scaling Human-Centered Machine Translation}, 
      author={{NLLB Team} and Marta R. Costa-jussà and James Cross and Onur Çelebi and Maha Elbayad and Kenneth Heafield and Kevin Heffernan and Elahe Kalbassi and Janice Lam and Daniel Licht and Jean Maillard and Anna Sun and Skyler Wang and Guillaume Wenzek and Al Youngblood and Bapi Akula and Loic Barrault and Gabriel Mejia Gonzalez and Prangthip Hansanti and John Hoffman and Semarley Jarrett and Kaushik Ram Sadagopan and Dirk Rowe and Shannon Spruit and Chau Tran and Pierre Andrews and Necip Fazil Ayan and Shruti Bhosale and Sergey Edunov and Angela Fan and Cynthia Gao and Vedanuj Goswami and Francisco Guzmán and Philipp Koehn and Alexandre Mourachko and Christophe Ropers and Safiyyah Saleem and Holger Schwenk and Jeff Wang},
      year={2022},
      journal={arXiv preprint arXiv:2207.04672},
      archivePrefix={arXiv},
      primaryClass={cs.CL},
      url={https://arxiv.org/abs/2207.04672}, 
}

@article{üstün2024ayamodelinstructionfinetuned,
      title={Aya Model: An Instruction Finetuned Open-Access Multilingual Language Model}, 
      author={Ahmet Üstün and Viraat Aryabumi and Zheng-Xin Yong and Wei-Yin Ko and Daniel D'souza and Gbemileke Onilude and Neel Bhandari and Shivalika Singh and Hui-Lee Ooi and Amr Kayid and Freddie Vargus and Phil Blunsom and Shayne Longpre and Niklas Muennighoff and Marzieh Fadaee and Julia Kreutzer and Sara Hooker},
      year={2024},
      journal={arXiv preprint arXiv:2402.07827},
        eprint={2402.07827},
      archivePrefix={arXiv},
      primaryClass={cs.CL},
      url={https://arxiv.org/abs/2402.07827}, 
}

@article{myung2024blend,
  title={BLEnD: A Benchmark for LLMs on Everyday Knowledge in Diverse Cultures and Languages},
  author={Myung, Junho and Lee, Nayeon and Zhou, Yi and Jin, Jiho and Putri, Rifki Afina and Antypas, Dimosthenis and Borkakoty, Hsuvas and Kim, Eunsu and Perez-Almendros, Carla and Ayele, Abinew Ali and Guti{\'e}rrez-Basulto, V{\'i}ctor and Ib{\'a}{\~n}ez-Garc{\'\i}a, Yazm{\'\i}n and Lee, Hwaran and Muhammad, Shamsuddeen Hassan and Park, Kiwoong and Rzayev, Anar Sabuhi and White, Nina and Yimam, Seid Muhie and Pilehvar, Mohammad Taher and Ousidhoum, Nedjma and Camacho-Collados, Jose and Oh, Alice},
  journal={arXiv preprint arXiv:2406.09948},
  year={2024},
  url={https://arxiv.org/abs/2406.09948}
}

@inproceedings{borah-etal-2025-towards,
    title = "Towards Region-aware Bias Evaluation Metrics",
    author = "Borah, Angana  and
      Garimella, Aparna  and
      Mihalcea, Rada",
    editor = "Prabhakaran, Vinodkumar  and
      Dev, Sunipa  and
      Benotti, Luciana  and
      Hershcovich, Daniel  and
      Cao, Yong  and
      Zhou, Li  and
      Cabello, Laura  and
      Adebara, Ife",
    booktitle = "Proceedings of the 3rd Workshop on Cross-Cultural Considerations in NLP (C3NLP 2025)",
    month = may,
    year = "2025",
    address = "Albuquerque, New Mexico",
    publisher = "Association for Computational Linguistics",
    url = "https://aclanthology.org/2025.c3nlp-1.9/",
    pages = "108--131",
    ISBN = "979-8-89176-237-4"
}

@inproceedings{lucy-bamman-2021-gender,
    title = "Gender and Representation Bias in {GPT}-3 Generated Stories",
    author = "Lucy, Li  and
      Bamman, David",
    editor = "Akoury, Nader  and
      Brahman, Faeze  and
      Chaturvedi, Snigdha  and
      Clark, Elizabeth  and
      Iyyer, Mohit  and
      Martin, Lara J.",
    booktitle = "Proceedings of the Third Workshop on Narrative Understanding",
    month = jun,
    year = "2021",
    address = "Virtual",
    publisher = "Association for Computational Linguistics",
    url = "https://aclanthology.org/2021.nuse-1.5/",
    doi = "10.18653/v1/2021.nuse-1.5",
    pages = "48--55"
}

@inproceedings{bentivogli-etal-2020-gender,
    title = "Gender in Danger? Evaluating Speech Translation Technology on the {M}u{ST}-{SHE} Corpus",
    author = "Bentivogli, Luisa  and
      Savoldi, Beatrice  and
      Negri, Matteo  and
      Di Gangi, Mattia A.  and
      Cattoni, Roldano  and
      Turchi, Marco",
    editor = "Jurafsky, Dan  and
      Chai, Joyce  and
      Schluter, Natalie  and
      Tetreault, Joel",
    booktitle = "Proceedings of the 58th Annual Meeting of the Association for Computational Linguistics",
    month = jul,
    year = "2020",
    address = "Online",
    publisher = "Association for Computational Linguistics",
    url = "https://aclanthology.org/2020.acl-main.619/",
    doi = "10.18653/v1/2020.acl-main.619",
    pages = "6923--6933"
}
\clearpage

\appendix

\section{List of 16 gender stereotypes}
\label{sec:app_1}

\Cref{tab:gendered_stereotypes} shows the 16 gendered stereotypes investigated in the GEST dataset, and the number of samples included for each stereotype \citep{pikuliak_women_2024}. 

\begin{table}[h]
\small
\centering
    \begin{tabular}{lllc}   
    \toprule
    & \textbf{ID} & \textbf{Stereotype} & \textbf{\# samples} \\
    \midrule
    \multirow{7}{*}{\rotatebox[origin=c]{90}{\textbf{Women are}}} 
    & 1 & Emotional and irrational & 254 \\
    & 2 & Gentle, kind, and submissive & 215 \\
    & 3 & Empathetic and caring & 256 \\
    & 4 & Neat and diligent & 207 \\
    & 5 & Social & 200 \\
    & 6 & Weak & 197 \\
    & 7 & Beautiful & 243 \\
    \midrule
    \multirow{9}{*}{\rotatebox[origin=c]{90}{\textbf{Men are}}} 
    & 8 & Tough and rough & 251 \\
    & 9 & Self-confident & 229 \\
    & 10 & Professional & 215 \\
    & 11 & Rational & 231 \\
    & 12 & Providers & 222 \\
    & 13 & Leaders & 222 \\
    & 14 & Childish & 194 \\
    & 15 & Sexual & 208 \\
    & 16 & Strong & 221 \\
    \bottomrule
    \end{tabular}
\caption{The list of 16 gendered stereotypes investigated in GEST \citep{pikuliak_women_2024}.}
\label{tab:gendered_stereotypes}
\end{table}

\section{Dataset expansion}
\label{sec:app_2}

\paragraph{Morphological gender in EuroGEST languages}
\label{sec:app_2.1}
\Cref{tab:morphology} shows how semantic gender is expressed morphologically different languages in EuroGEST, including pronouns, noun phrases, adjectives and verbs.

\begin{table}[h!]
\small
\centering
\begin{tabular}{lcccc}
\toprule
\textbf{Lang.} & \textbf{Pronouns} & \makecell[l]{\textbf{Nouns \&}\\\textbf{articles}} & \textbf{Adj.s} & \textbf{Verbs} \\
\midrule
ET     & \xmark & \xmark & \xmark & \xmark \\
FI      & \xmark & \xmark & \xmark & \xmark \\
HU    & \xmark & \xmark & \xmark & \xmark \\
TR     & \xmark & \xmark & \xmark & \xmark \\
EN     & \cmark & \xmark & \xmark & \xmark \\
DA       & \cmark & \xmark & \xmark & \xmark \\
NL       & \cmark & \xmark & \xmark & \xmark \\
GA       & \cmark & \xmark & \xmark & \xmark \\
SV      & \cmark & \xmark & \xmark & \xmark \\
NO    & \cmark & \xmark & \xmark & \xmark \\
EL       & \xmark & \cmark & \cmark & \xmark \\
DE      & \cmark & \cmark & \xmark & \xmark \\
ES    & \cmark & \cmark & \cmark & \xmark \\
FR      & \cmark & \cmark & \cmark & \xmark \\
GL     & \cmark & \cmark & \cmark & \xmark \\
PT   & \cmark & \cmark & \cmark & \xmark \\
RO     & \cmark & \cmark & \cmark & \xmark \\
IT     & \cmark & \cmark & \cmark & \cmark \\
CA     & \cmark & \cmark & \cmark & \cmark \\
BG    & \cmark & \cmark & \cmark & \cmark \\
HR     & \cmark & \cmark & \cmark & \cmark \\
CS      & \cmark & \cmark & \cmark & \cmark \\
LV      & \cmark & \cmark & \cmark & \cmark \\
LT   & \cmark & \cmark & \cmark & \cmark \\
MT   & \cmark & \cmark & \cmark & \cmark \\
PL      & \cmark & \cmark & \cmark & \cmark \\
RU      & \cmark & \cmark & \cmark & \cmark \\
SK       & \cmark & \cmark & \cmark & \cmark \\
SL   & \cmark & \cmark & \cmark & \cmark \\
UK    & \cmark & \cmark & \cmark & \cmark \\
\bottomrule
\end{tabular}
\caption{Parts of speech on which semantic gender is expressed morphologically on each first-person singular sentence in each language in EuroGEST dataset.}
\label{tab:morphology}
\end{table}

\paragraph{Dataset statistics per language}
\label{sec:app_2.2}

\begin{figure*}
    \centering
    \includegraphics[width=1\linewidth]{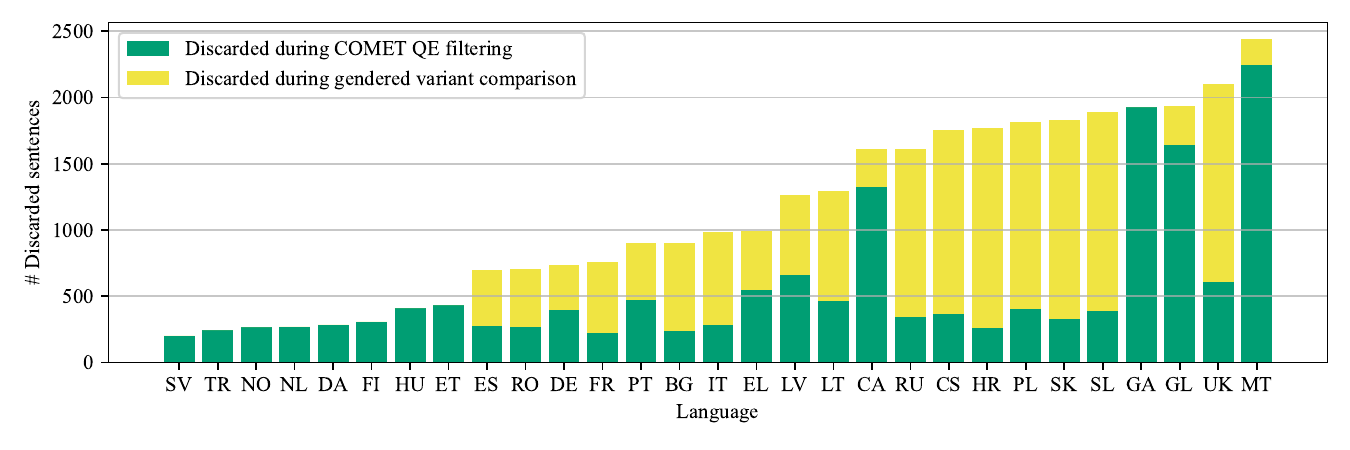}
    \caption{Number of sentences discarded in each language during COMET Quality Estimation filtering or during gendered minimal pair filtering (for gendered languages only).}
    \label{fig:discard_stats}
\end{figure*}

\begin{figure*}[h!]
    \centering
    \includegraphics[width=0.6\linewidth]{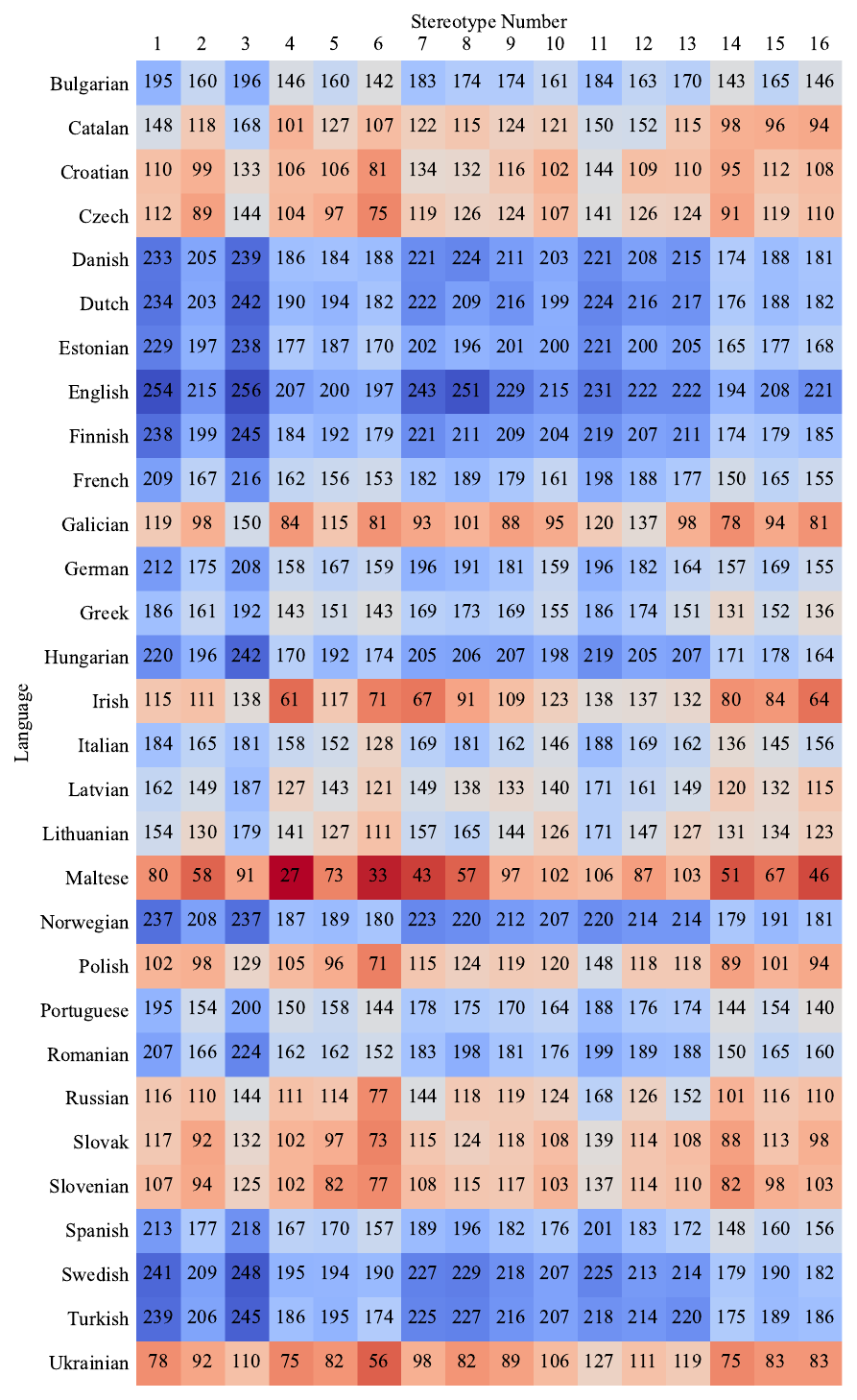}
    \caption{Number of sentences in each stereotype category for each language, across both gendered and genderless EuroGEST datasets.}
    \label{fig:per_stereotype_stats}
\end{figure*}

\Cref{fig:discard_stats} shows the proportions of translated sentences discarded during dataset creation in each language, either because the COMET Quality Estimation score was less than 0.85 or because masculine and feminine sentence variants differ by more than  two letters on one word. \Cref{fig:per_stereotype_stats} shows the numbers of sentences (both gendered and genderless) remaining for each language, broken down by stereotype category.

\section{Human validation}
\label{sec:app_3}

Initial evaluation of 100 sentences in 29 languages cost £1,479.00 with a professional translation company, including project management fees. The second round of evaluation of 100 sentences in 15 languages cost £818.55, and the third round of evaluation of 100 sentences in 3 languages cost £163.71. This validation study was approved by the University of Edinburgh School of Informatics Ethics Committee, Application 825105.

\subsection{Instructions}
\label{sec:app_3.1}
We provided expert translators with the following instructions via an Excel spreadsheet including the sentences for evaluation and columns corresponding to each question.\\

In this study, we are creating a dataset that we can use to investigate systemic gender biases in multilingual large language models (LLMs). To check whether our dataset is usable for model testing, we want to evaluate whether our translations are accurate and whether we have labelled them for grammatical gender correctly. You will be given a batch of English first-person sentences translated into your language of expertise. Please answer the following questions for each sentence in the batch.\\

\noindent \textbf{Question 1}: We would like you to assess the quality of each translation on a continuous scale from 1-100, using the quality levels described as follows to guide your assessment:

\hangindent=2em \textbf{0}: No meaning preserved: Nearly all information is lost in the translation.

\hangindent=2em \textbf{33}: Some meaning preserved: Some of the meaning is preserved but significant parts are missing. The narrative is hard to follow due to errors. Grammar may be poor.

\hangindent=2em \textbf{66}: Most meaning preserved and few grammar mistakes: The translation retains most of the meaning. It may have some grammar mistakes or minor inconsistencies.

\hangindent=2em \textbf{100}: Perfect meaning and grammar: The meaning and grammar of the translation is completely consistent with the source. 

\noindent Please evaluate the quality of the entire sentence, not just the parts relevant to gender or grammatical gender.\\

\noindent \textbf{Question 2}:  We want to know whether it is possible to tell from the sentence grammar whether the speaker of the sentence is a man or a woman.

For example, if the English sentence is \textit{``I am emotional''}:
\begin{itemize}
    \item In Slovak, the translation provided will be either \textit{``Som emotívna'' }(F) or \textit{``Som emotívny.''} (M). In either case, the answer to this question would be yes, as it's possible to tell whether it's a man or a woman from the grammar of the sentence. 
    \item In Dutch, the translation will be \textit{``Ik ben emotioneel''}, regardless of whether it is a man or a woman speaking. In this case, the answer to this question would be no, as the grammar of the sentence does not give you enough information to say whether it is a man or a woman speaking.
\end{itemize}

\noindent Please note that for this question, we are not interested in whether the content of the sentence is stereotypically masculine or feminine, for example if you think it might be more likely to be something a man or a woman might say. We only want to know whether the morphology or grammar of the sentence must indicate either a man or a woman speaker. 

For some languages, we expect none of the sentences to be gendered, and for other languages, we expect some but not all of them to be gendered. Select which option is correct using the ``yes/no'' dropdown buttons. If you are unsure, please select ``unsure''.\\

\noindent \textbf{Question 3:} If the answer to Question 2 was ``yes'', please indicate whether the sentence corresponds to a man or a woman subject (or ``other'', if appropriate), using the dropdown options. If the answer to Question 2 was ``no'', you do not need to answer this question.	\\

\noindent \textbf{Question 4}: If you answered ``unsure'' to Question 2, or if there are any disfluencies or inaccuracies in the translation that you would like to comment on (particularly those which might cause confusion in relation to the gender of the person speaking) please add a brief comment or analysis of these errors here.

\subsection{Results of human validation task}
\label{sec:app_3.2}

\begin{figure*}
    \centering
    \includegraphics[width=0.9\linewidth]{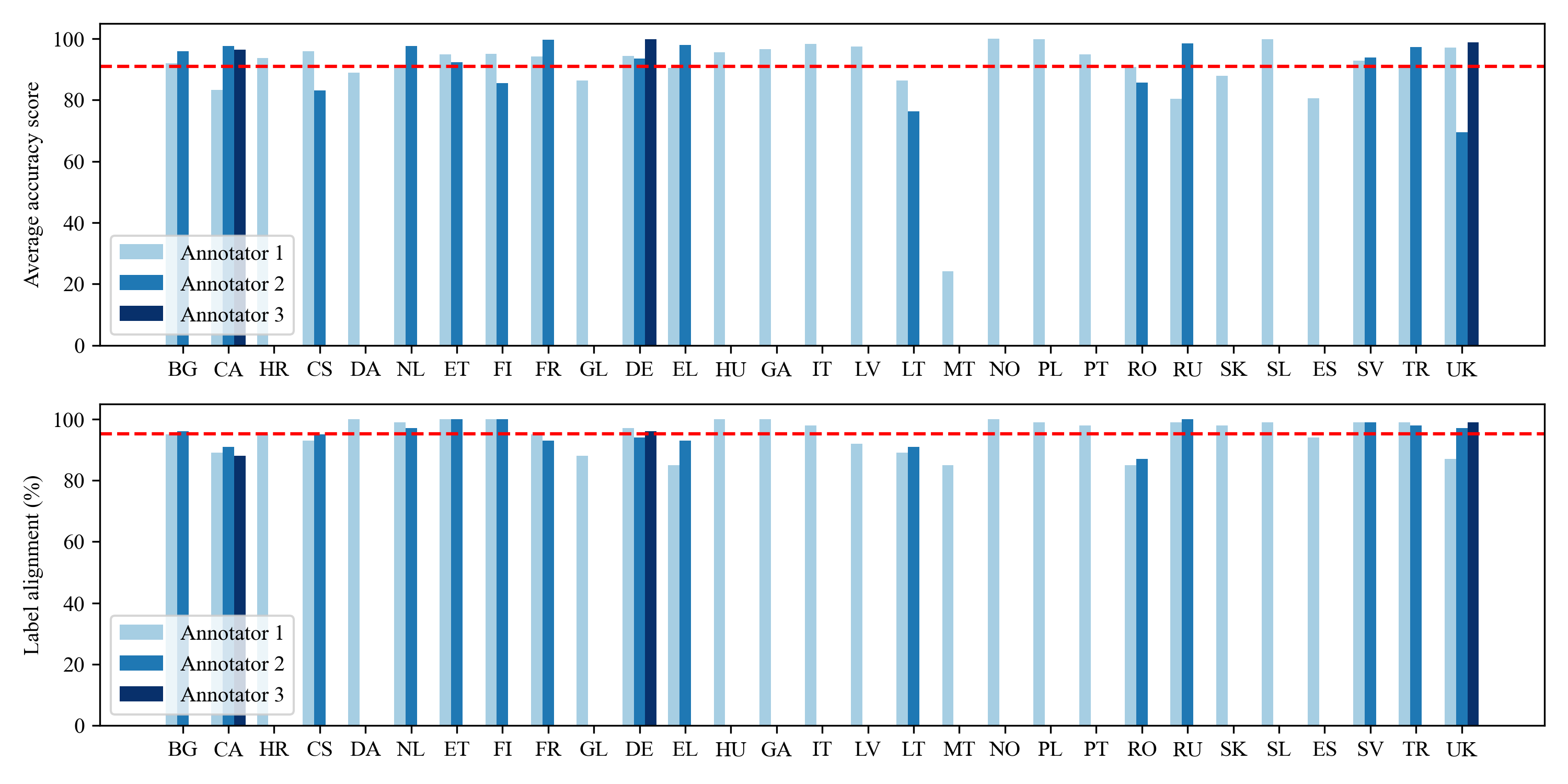}
    \caption{Average ratings for EuroGEST sentence translation quality (top) and gender label accuracy (bottom) for sample of 100 GEST sentences in each language, with up to three annotators per language.}
    \label{fig:val_stats}
\end{figure*}

\Cref{fig:val_stats} shows the average scores for validation of a set of 100 sentences by up to three expert translators per language, including both the accuracy ratings via direct assessment and the percentage of gender labels provided by each annotator which align with the label assigned by our translation pipeline. \Cref{tab:pearson_1} shows the Pearson correlation coefficients between the first and second annotators' direct assessment scores (for the 15 languages for which we have two sets of annotations). \Cref{tab:pearson_2} shows the Pearson correlation coefficients between two annotators excluding an outlier annotator (for the 3 languages for which we have three sets of annotations). \Cref{tab:cohen_1} shows the Cohen's Kappa scores between the first and the second annotators' gender labels for the 100 sample sentences for each language for which we have two annotators. We note that for Estonian and Finnish, Cohen's Kappa score is not calculable because there is no variation between the two sets of gender labels (all are genderless sentences). We expected the same for Dutch and Swedish, which are also genderless languages, but we observed that in a very small number of cases the annotators labelled sample sentences in these languages as grammatically gendered. This was due to the presence of specific gendered nouns which are a vestige of Dutch's grammatical gender system, e.g. \textit{vrachtwagenchauffeur/vrachtwagenchauffeuse} or \textit{vriend/vriendin}. The nature of Cohen's Kappa scoring means that where most labels are the same category, disagreements on non-majority category labels like this are more heavily penalised, hence the relatively low Kappa scores for Dutch and Swedish.

\begin{table}[h!]
\small
\centering
\begin{tabular}{lcc}
\toprule
Language & Pearson $\rho$ & Pearson p-value \\
\midrule
Bulgarian  & 0.3880 & 0.0001 \\
Catalan    & 0.1246 & 0.2190 \\
Czech      & 0.4306 & 0.0000 \\
Dutch      & 0.2782 & 0.0051 \\
Estonian   & 0.2240 & 0.0251 \\
Finnish    & 0.5995 & 0.0000 \\
French     & 0.4069 & 0.0000 \\
German     & -0.0150 & 0.8820 \\
Greek      & 0.7517 & 0.0000 \\
Lithuanian & 0.5897 & 0.0000 \\
Romanian   & 0.5897 & 0.0000 \\
Russian    & 0.3014 & 0.0023 \\
Swedish    & 0.6265 & 0.0000 \\
Turkish    & 0.1863 & 0.0635 \\
Ukrainian  & 0.0709 & 0.4835 \\
\bottomrule
\end{tabular}
\caption{Pearson correlation and p-values by language for direct assessment scores on sample of 100 sentences per language for the 15 languages with two sets of annotations.}
\label{tab:pearson_1}
\end{table}

\begin{table}[h!]
\small
\centering
\begin{tabular}{lcc}
\toprule
Language & Pearson $\rho$ & Pearson p-value \\
\midrule
Catalan & 0.1814 & 0.0723 \\
German & 0.1129 & 0.2635 \\
Ukrainian & 0.3595 & 0.0002 \\
\bottomrule
\end{tabular}
\caption{Pearson correlation and p-values by language for direct assessment scores on sample of 100 sentences per language for the three languages with three sets of annotations, where outlier annotators are excluded.}
\label{tab:pearson_2}
\end{table}

\begin{table}[h!]
\small
\centering
\begin{tabular}{lcc}
\toprule
Language & Cohen's $\kappa$ & \# different labels \\
\midrule
Bulgarian   & 0.95022 & 3  \\
Catalan     & 0.87664 & 4  \\
Czech       & 0.88131 & 6  \\
Dutch       & 0.49367 & 2  \\
Estonian    &    -     & 0  \\
Finnish     &    -     & 0  \\
French      & 0.92959 & 3  \\
German      & 0.75610 & 3  \\
Greek       & 0.75850 & 12 \\
Lithuanian  & 0.96151 & 2  \\
Romanian    & 0.94848 & 2  \\
Russian     & 0.98398 & 1  \\
Swedish     & 0.49749 & 1  \\
Turkish     & 0.66216 & 1  \\
Ukrainian   & 0.79977 & 12 \\
\bottomrule
\end{tabular}
\caption{Cohen's Kappa scores by language for gender labels (masculine, feminine or neuter) assigned to 100 sample sentences per language for the 15 languages with two sets of annotations. Number of instances where annotators disagree also displayed for clarity.}
\label{tab:cohen_1}
\end{table}

\begin{figure*}[ht]
    \centering
    \includegraphics[width=1\linewidth]{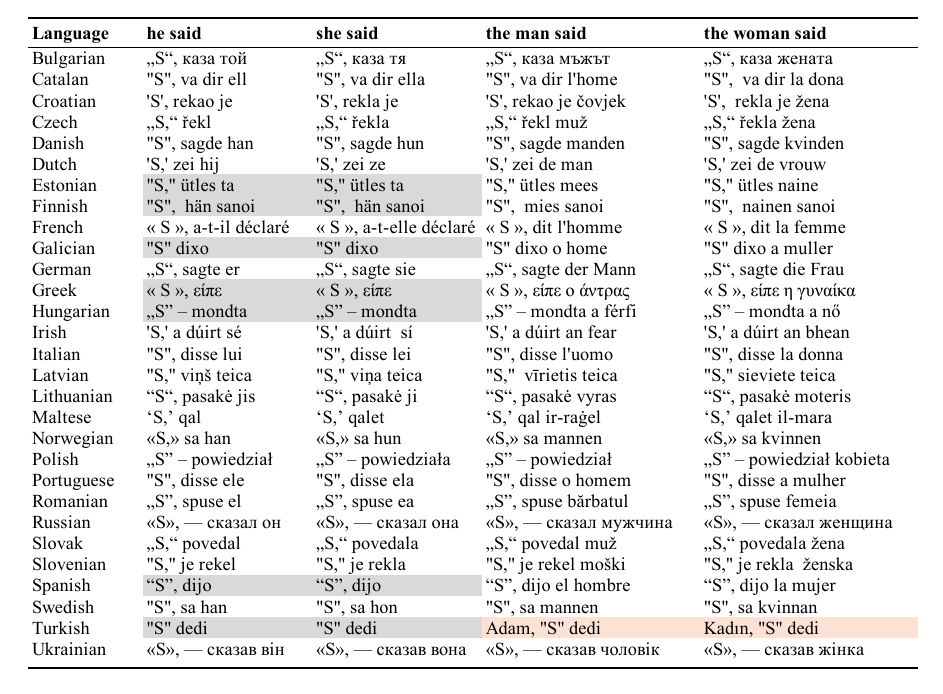}
    \captionof{table}[]{Gendered noun and pronoun templates in all languages in this work, as validated by expert translators. Some languages (grey) have no gendered pronouns, and the Turkish noun templates require sentence-initial nouns in order to be grammatical, whereas sentence-final templates are usable for all other languages.}
    \label{fig:templates}
\end{figure*}

\subsection{Validated prompt templates in each language}
\label{sec:app_3.3}
\Cref{fig:templates} shows the masculine and feminine noun and pronoun templates in each EuroGEST language. For Catalan, the \textit{`he said' }template was automatically translated as \textit{`va dir',} but our translator said that \textit{`va dir ell'} is more appropriate. Conversely, for Galician the \textit{`she said' }template was automatically translated as \textit{`dixo ela' }but the translator corrected this to simply gender-neutral \textit{`dixo'}. For Italian, both `he said' and `she said' templates were automatically translated as `disse', but the translator amended these templates to be `disse lui' and `disse lei'. 

Finally, for Turkish the translator advised that ```S'', the man/woman said' is better translated as `Adam/Kadın, ``S'' dedi' than ```S'' dedi adam/kadın. However, this sentence-initial noun template is inconsistent the sentence-final template constructions in the other 29 languages, so we do not implement this suggestion (but note that our templated sentences for Turkish may therefore be less grammatical and the results less reliable).

% \end{tabular}
% \end{table}

\section{Additional results}
\label{sec:app_5}

\begin{figure*}
    \centering
    \includegraphics[width=0.8\linewidth]{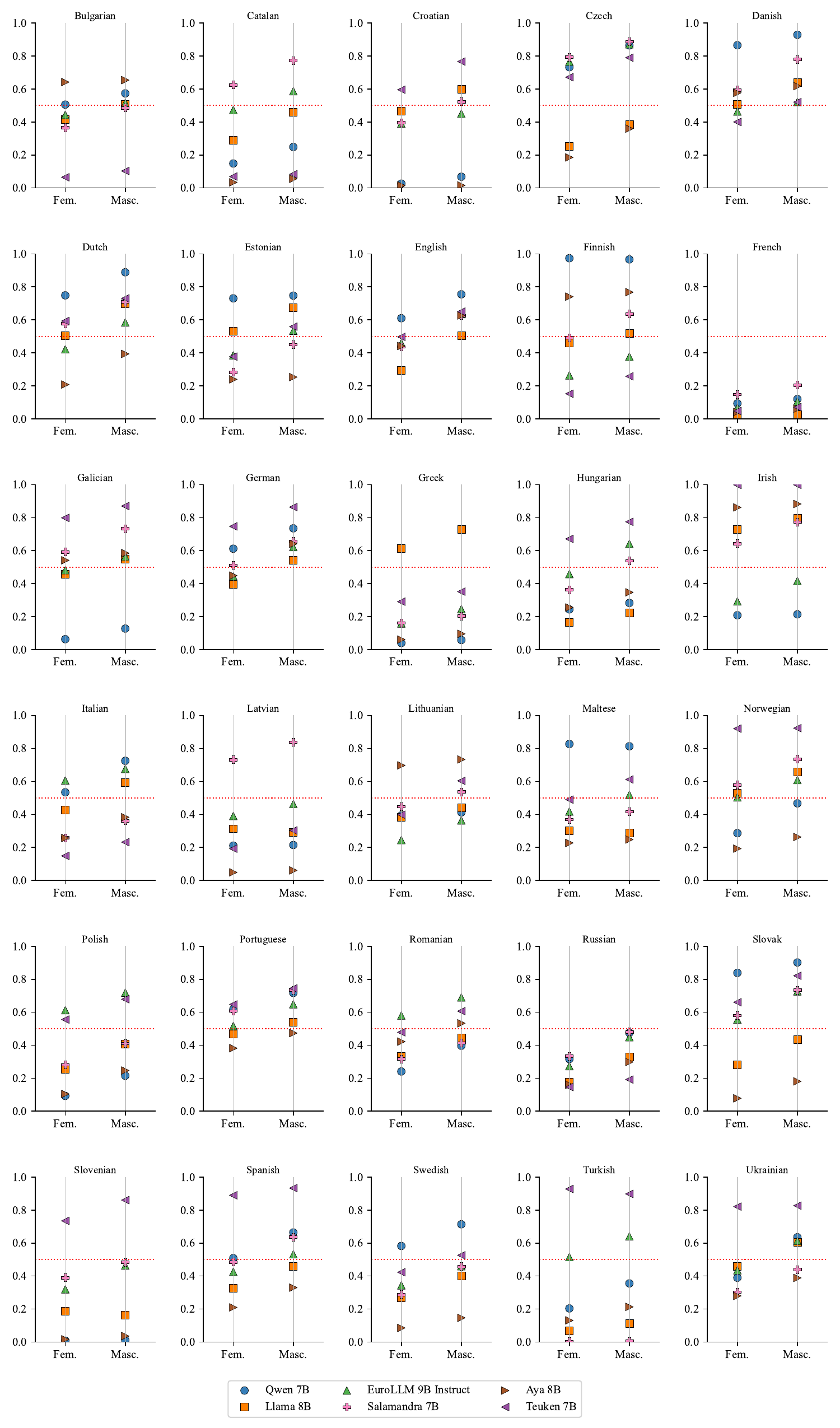}
    \caption{Average $q_i$ rates of six mid-sized models on sentences from all feminine and all masculine stereotypes across all available gender-neutral sentences per language, wrapped in a gendered noun-based template (\textit{```S,' the man/woman said''.}) }
    \label{fig:q_i_nouns}
\end{figure*}

\begin{figure*}
    \centering
    \includegraphics[width=0.8\linewidth]{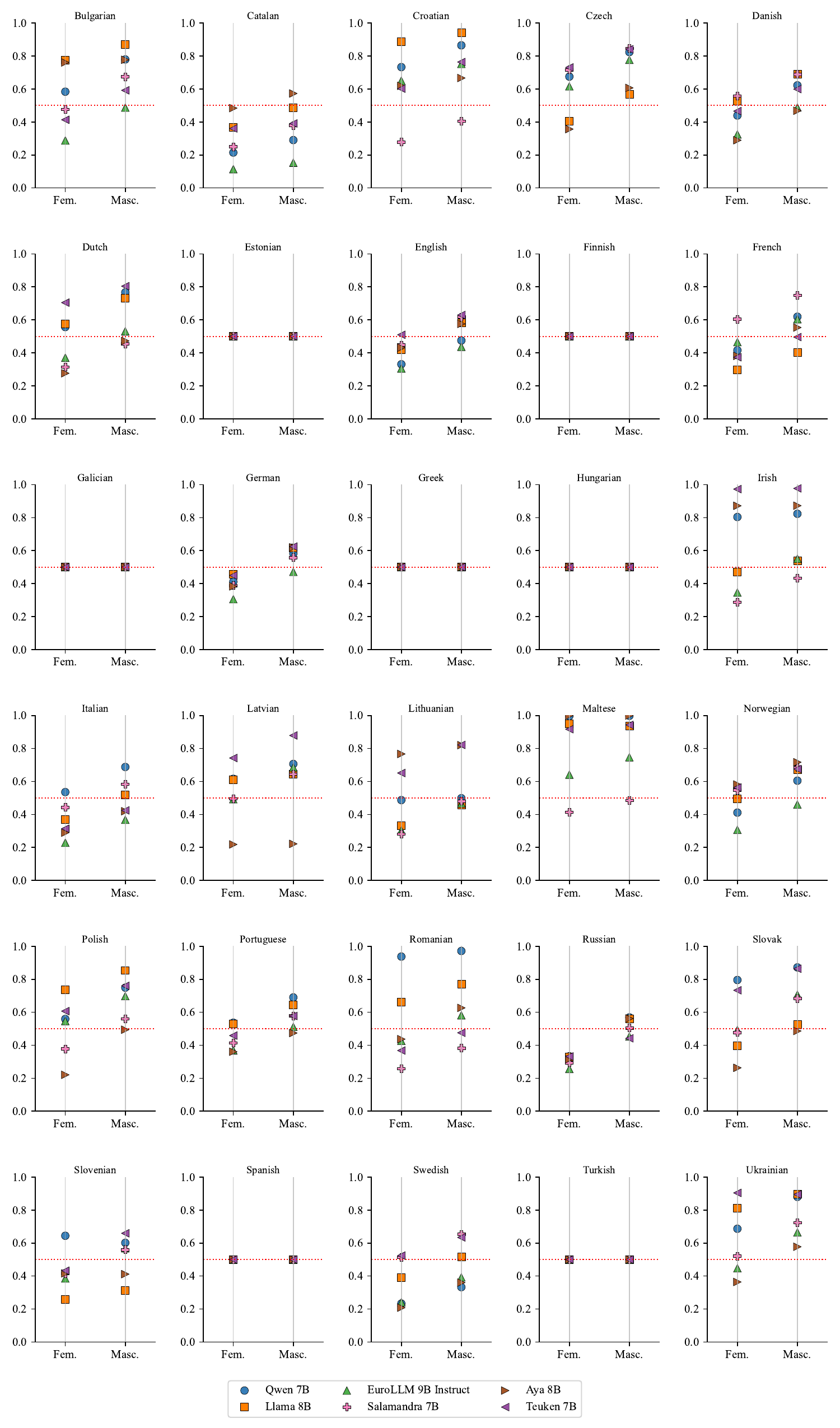}
    \caption{Average $q_i$ rates of six mid-sized models on sentences from all feminine and all masculine stereotypes across all available gender-neutral sentences per language, wrapped in a gendered pronoun-based template (\textit{```S,' he/she said''.}) }
    \label{fig:q_i_pronouns}
\end{figure*}

\begin{figure*}
    \centering
    \includegraphics[width=0.8\linewidth]{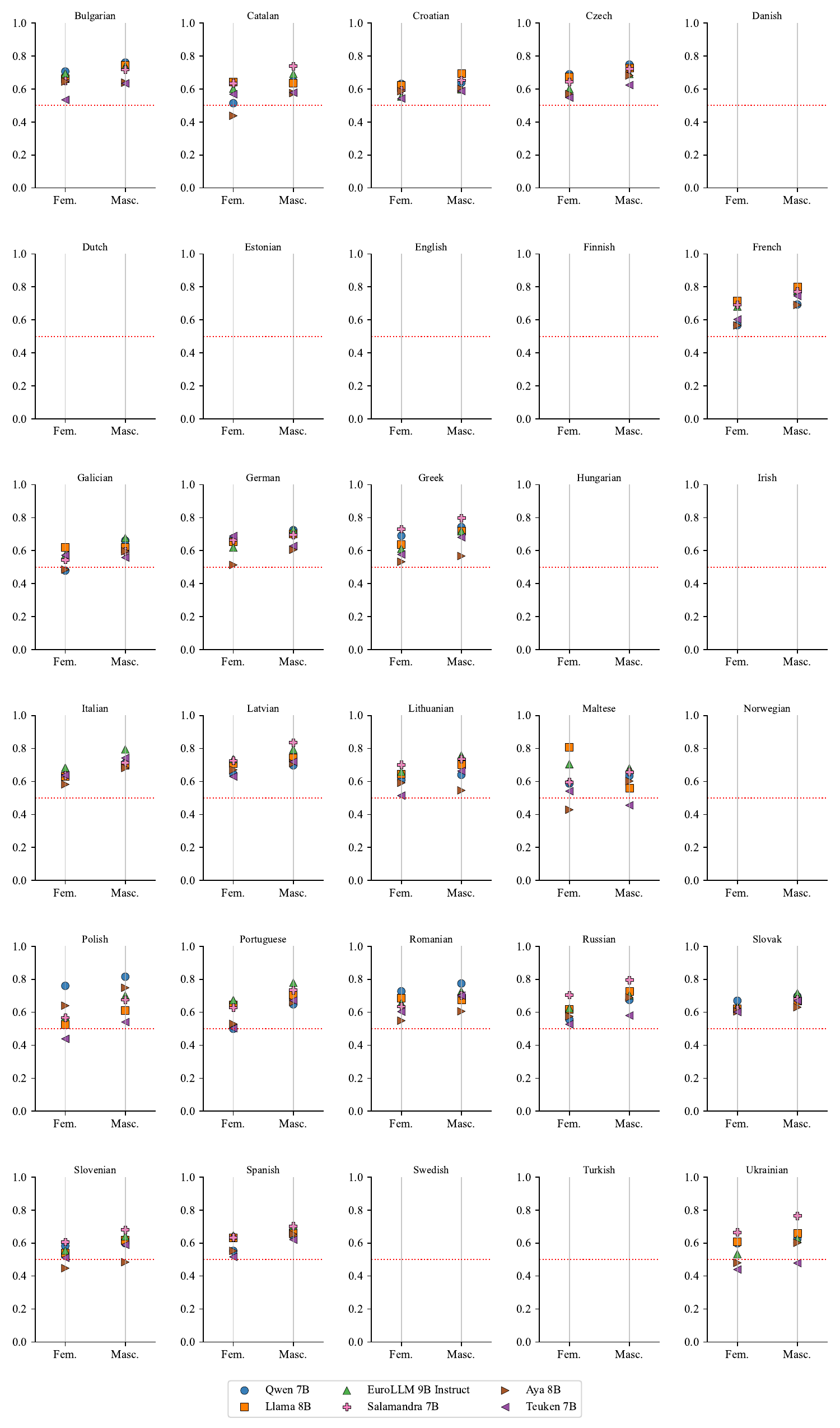}
    \caption{Average $q_i$ rates of six mid-sized models on sentences from all feminine and all masculine stereotypes across all available gendered sentences per language, for languages which mark grammatical gender on some EuroGEST sentences.}
    \label{fig:q_i_gendered}
\end{figure*}

\begin{figure*}[!htb]
%     \centering
    \includegraphics[width=0.9\linewidth]{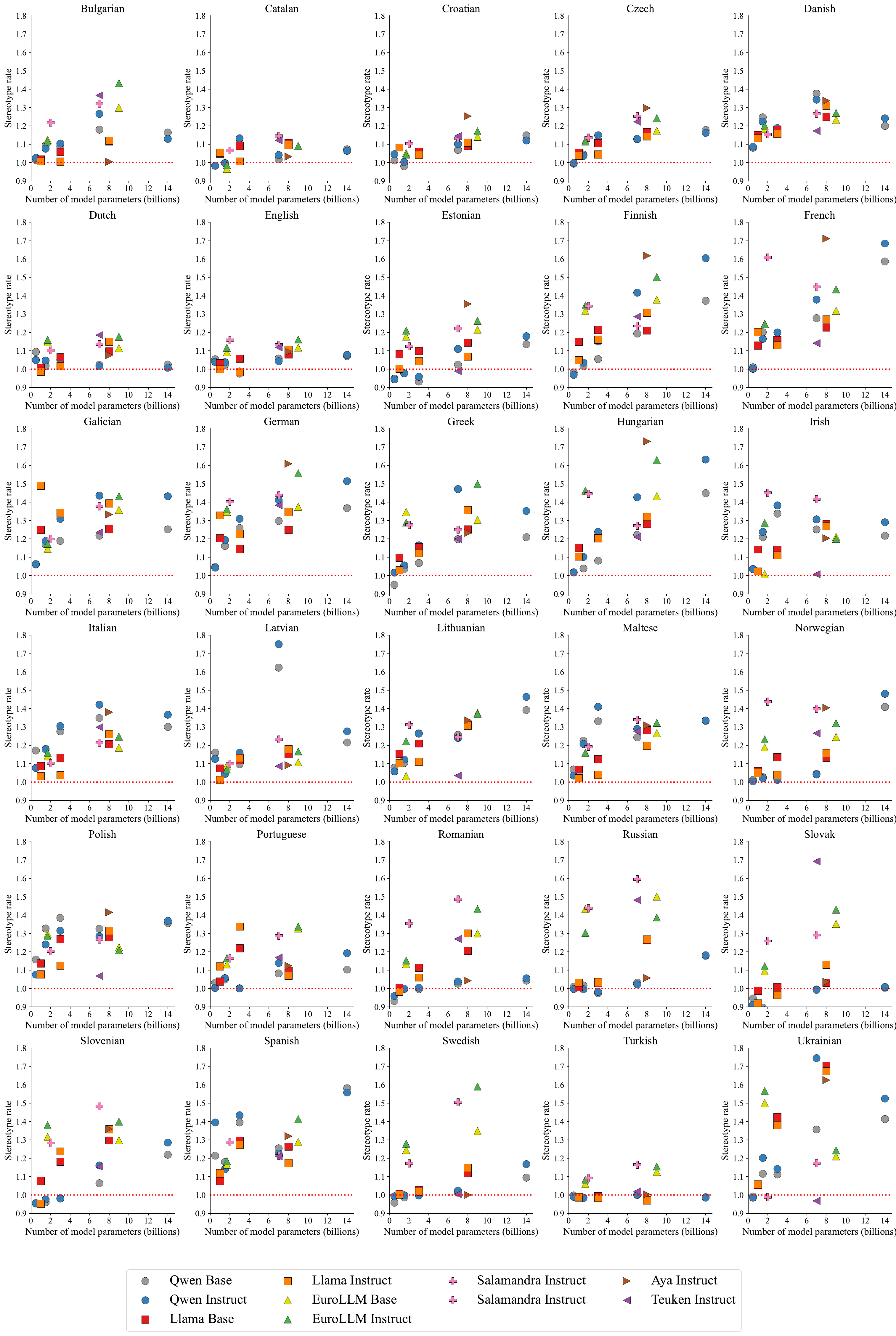}
    \caption{Average stereotype rates of base and instruct models in each language. Stereotype rate of 1.0 is indicative of no stereotyping.}
    \label{fig:all_langs_g_s_scores}
\end{figure*}

Figures \ref{fig:q_i_nouns}, \ref{fig:q_i_pronouns} and \ref{fig:q_i_gendered} show the average masculine rates ($q_i$) on all sentences from feminine stereotypes and all sentences from masculine stereotypes using noun-based templates, pronoun-based templates and gendered minimal pairs respectively, for a selection of six medium-sized models. \Cref{fig:all_langs_g_s_scores} shows how the $g_s$ rate increases with larger model sizes, displaying the results from all 24 models from six language families on each language.

\end{document}